%% file: main.tex
\documentclass[research]{fcs}
\usepackage{layout}
\usepackage{subfig}
\usepackage{multirow}
\usepackage{hyperref}
\usepackage{framed}
\usepackage{colortbl}
\usepackage{xcolor}
\usepackage{color}
\usepackage{pifont}

\def\benchmarkname{GraphInstruct}
\def\modelname{GraphSolver}
\def\modelnamestep{GraphSolver+}

\definecolor{lightgreen}{RGB}{224,255,224}

\title{\benchmarkname{}: Empowering Large Language Models with Graph Understanding and Reasoning Capability}

\author[1]{Zihan~Luo}
\author[1]{Xiran~Song}
\author[1,+]{Hong~Huang}
\author[3]{Jianxun~Lian}
\author[2]{Chenhao~Zhang}
\author[2]{Jinqi~Jiang}
\author[3]{Xing~Xie}
\author[1]{Hai~Jin}
\address[1]{National Engineering Research Center for Big Data Technology and System, Services Computing Technology and System
Lab, Cluster and Grid Computing Lab, School of Computer Science and Technology, Huazhong University of Science and Technology, Wuhan 430074, China}
\address[2]{School of Computer Science and Technology, Huazhong University of Science and Technology, Wuhan 430074, China}
\address[3]{Microsoft Research Asia, Beijing 100190, China}

\corremail{honghuang@hust.edu.cn}
\fcssetup{
  received = {month dd, yyyy},
  accepted = {month dd, yyyy},
}

\begin{abstract}
Improving the general capabilities of \textit{Large Language Models} (LLMs) is an active research topic. 
As a common data structure in many real-world domains, understanding graph data is a crucial part of advancing general intelligence.
To this end, we propose a dynamic benchmark named \textbf{\benchmarkname} in this paper, which comprehensively includes 21 classical graph reasoning tasks, providing diverse graph generation pipelines and detailed intermediate reasoning steps for each sample. 
Based on \benchmarkname{}, we develop \modelname{} via efficient instruction-tuning, which demonstrates prominent graph understanding capability compared to other open-sourced LLMs. To further endow LLMs with multi-step graph reasoning capability, we propose a label-mask training strategy and build \modelnamestep{}, which leverages masked supervision on intermediate reasoning tokens to emphasize crucial node-identification signals. As one of the pioneering efforts to enhance the graph understanding and reasoning abilities of LLMs, extensive experiments have demonstrated the superiority of \modelname{} and \modelnamestep{} over other LLMs. We sincerely hope \benchmarkname{} will facilitate further research on applying LLMs to graph-structured data. Our code and data are released publicly at: \url{https://github.com/CGCL-codes/GraphInstruct}.
\end{abstract}

\begin{document}

\input{introduction}

\input{related_work}

\input{benchmark}

\input{model}

\input{experiment}

\input{reasoning}

\input{conclusion}

\input{limitations}

\section*{Acknowledgements}
This work was supported by the National Natural
Science Foundation of China (Grant No. 62172174).

\section*{Competing Interest}
The authors declare that they have no competing interests or financial conflicts to disclose.

\bibliographystyle{fcs}
\bibliography{ref.bib}

\begin{biography}{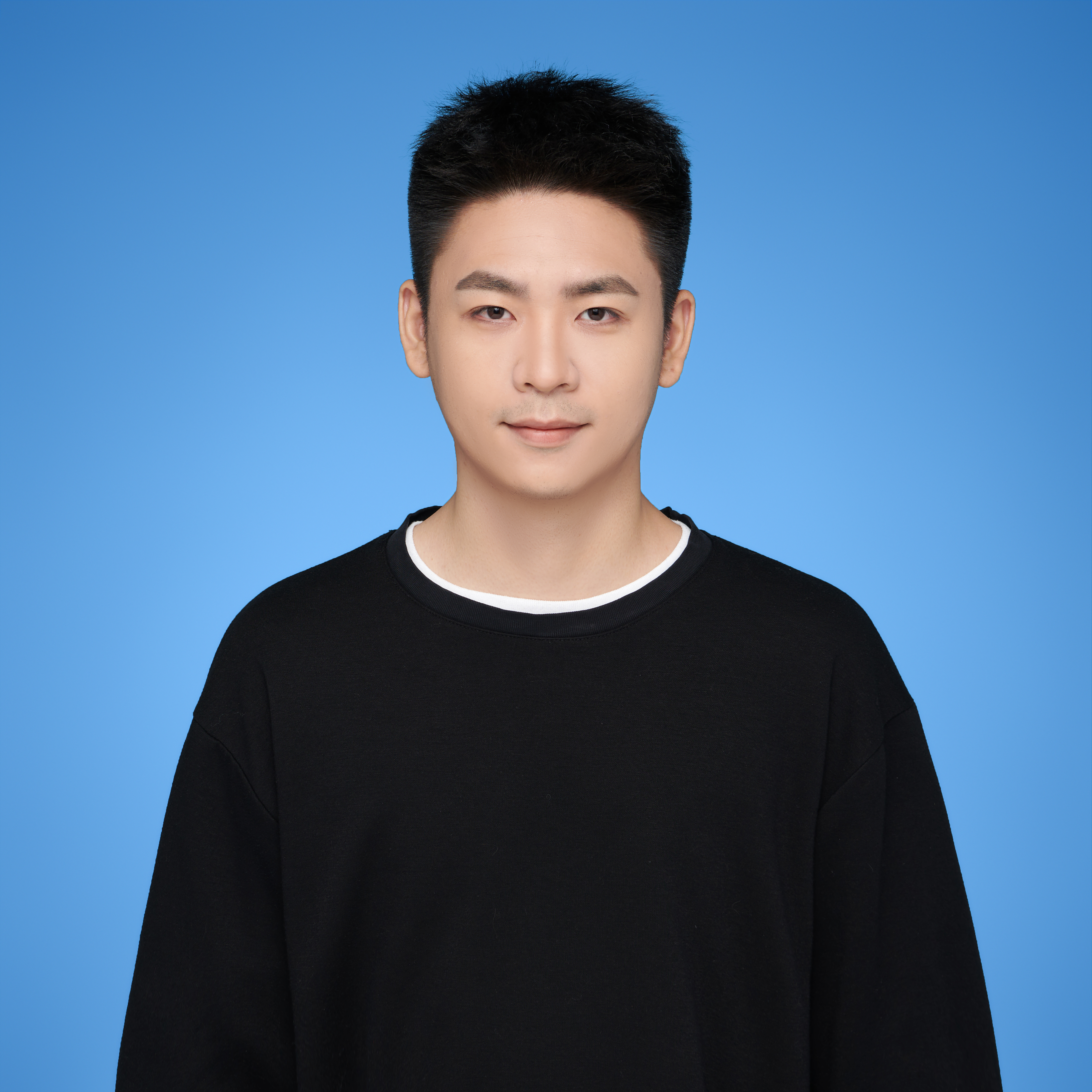}
{Zihan Luo} received the BEng degree from Huazhong University of Science and Technology, Wuhan, China, in 2020. He is currently working toward the PhD degree in the School of Computer Science at Huazhong University of Science and Technology. His research interests include large language models and graph representation learning.
\end{biography}

\begin{biography}{figure/author2.pdf}
{Xiran Song} received his MS degree from Huazhong University of Science and Technology, Wuhan, China, in 2024. He is currently pursuing a PhD in computer science at Mila, Quebec AI Institute, affiliated with Université de Montréal. His current research focuses on multi-modal large language models and embodied intelligence.
\end{biography}

\begin{biography}{figure/author3.pdf}
{Hong Huang} is an Associate Professor in Huazhong University of Science and Technology, China. She received her PhD in Computer Science from University of G\"{o}ttingen, Germany in 2016, and her M.E. degree in Electronic Engineering from Tsinghua University, Beijing, China in 2012. Her research interests lie in social network analysis, data mining and knowledge graph.
\end{biography}

\begin{biography}{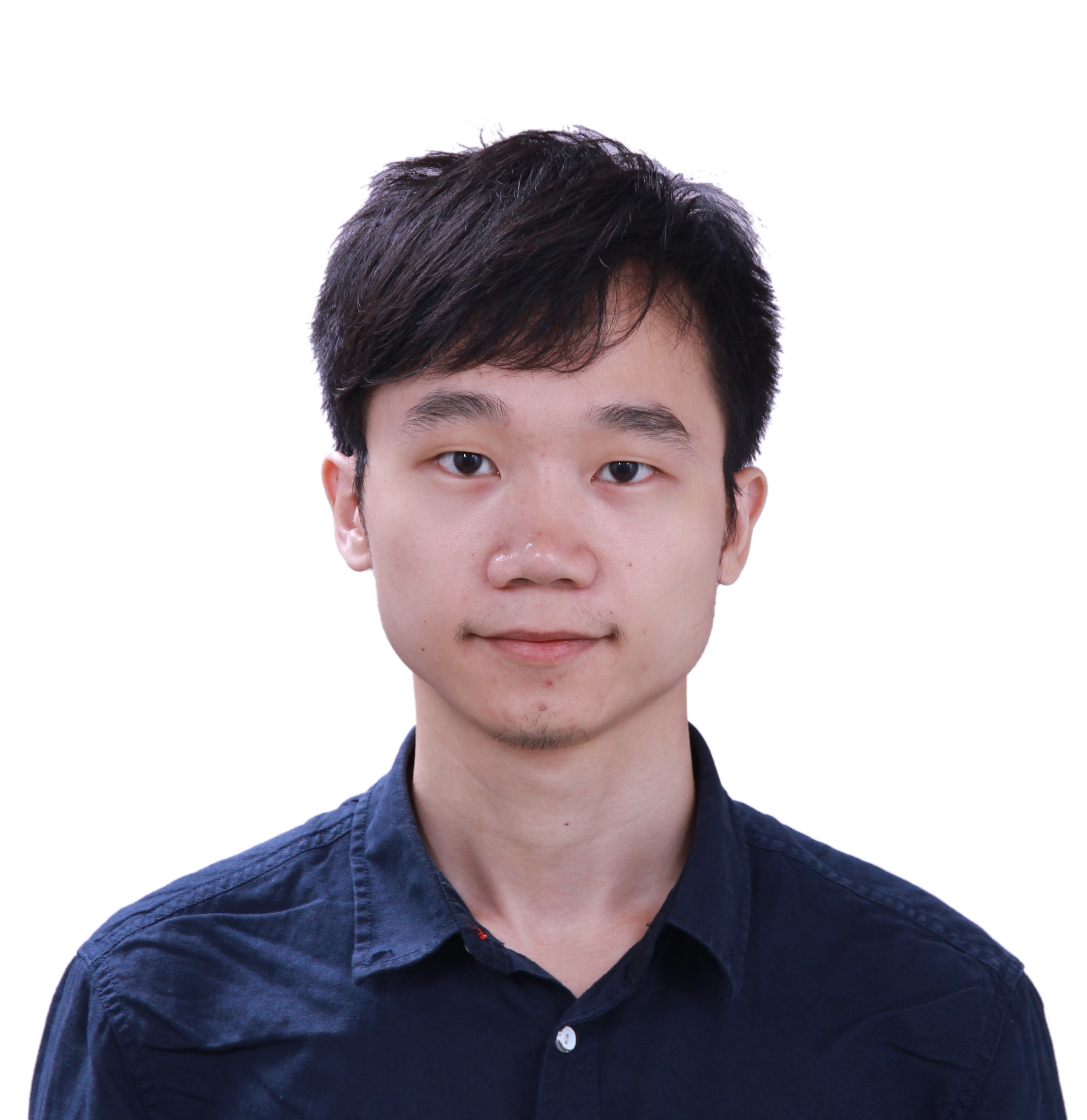}
{Jianxun Lian} is now a senior researcher at Microsoft Research Asia. He received his Ph.D. degree from University of Science and Technology of China in 2018. His research interests include recommender systems and deep learning techniques. He has published some academic papers on international conferences such as KDD, IJCAI, WWW, SIGIR and CIKM. He serves as a program committee member for several top conferences such as AAAI, WWW and IJCAI.
\end{biography}

\begin{biography}{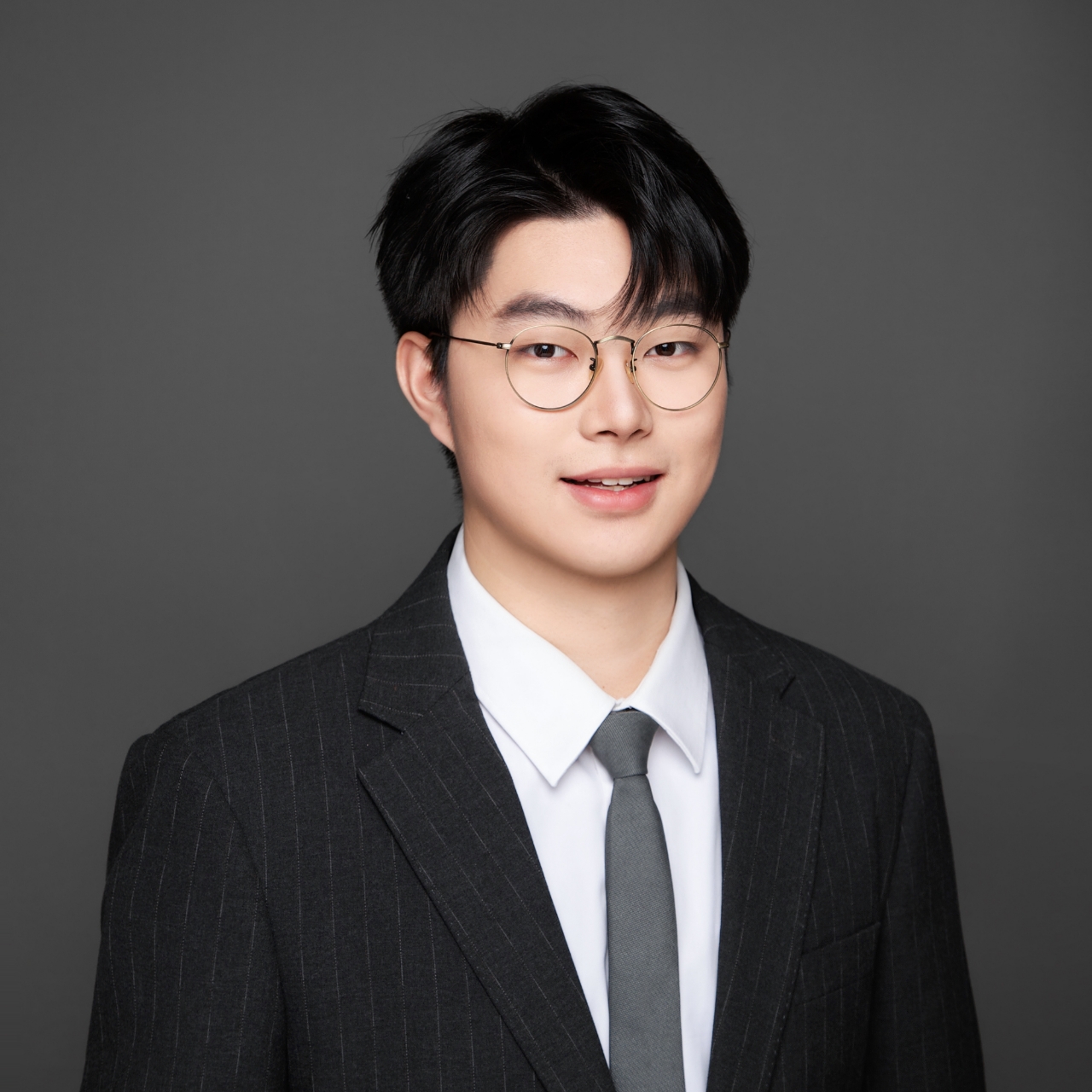}
{Chenhao Zhang} is an undergraduate student from Huazhong University of Science and Technology, China. He is currently pursuing the BS degree in Computer Science (2023 - 2027). His research interests include Vision-Language Reasoning and Emotional AI.
\end{biography}

\begin{biography}
{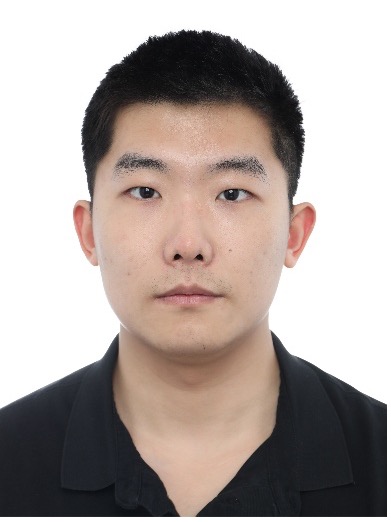} {Jinqi Jiang} is a senior undergraduate student at Huazhong University of Science and Technology, majoring in artificial intelligence. His research interests include large language models, efficient ai, and machine learning system.
\end{biography}

\begin{biography}{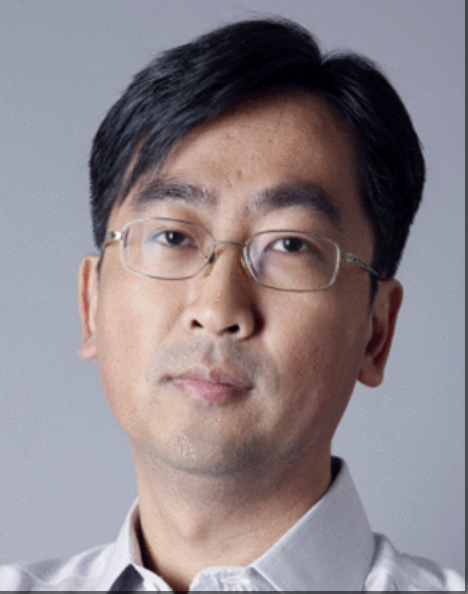}
Dr. Xing Xie is currently a senior principal research manager at Microsoft Research Asia, and a guest Ph.D. advisor at the University of Science and Technology of China. He received his B.S. and Ph.D. degrees in Computer Science from the University of Science and Technology of China in 1996 and 2001, respectively. He joined Microsoft Research Asia in July 2001, working on data mining, social computing and ubiquitous computing.
\end{biography}

\begin{biography}{figure/author8.pdf}
{Hai Jin} is a Chair Professor of computer science and engineering at Huazhong University of Science and Technology (HUST) in China. Jin received his PhD in computer engineering from HUST in 1994.
Jin is a Fellow of IEEE, Fellow of CCF, and a life member of the ACM. He has co-authored more than 20 books and published over 900 research papers. His research interests include computer architecture, parallel and distributed computing, big data processing, data storage, and system security.
\end{biography}

\end{document}

%% file: introduction.tex
\section{Introduction}

Recently, the rise of \textit{Large Language Models} (LLMs) has showcased their powerful generalization abilities in various domains such as mathematics \cite{t1,wizardmath}, code \cite{DBLP:journals/corr/abs-2306-08568,mle-bench}, and dialogue \cite{DBLP:conf/nips/Ouyang0JAWMZASR22,CPsyCoun}, and researchers are aspiring to leverage LLMs to achieve the capabilities of artificial general intelligence. With this goal in mind, many researchers have begun efforts to assist LLMs in understanding diverse forms of input data, including images \cite{DBLP:conf/nips/AlayracDLMBHLMM22,II-Bench}, audios \cite{DBLP:journals/corr/abs-2304-12995}, and others. However, this body of work has overlooked the capability of LLMs to comprehend data in the form of graphs. In fact, as a common data structure in the real world, graphs are prevalent in many scenarios like social networks \cite{xgcn}, fraud detection \cite{fraud}, and biological molecules \cite{bio,he2024integrating}. What's more, many logical reasoning tasks also fundamentally exist in the form of graphs \cite{anonymous2024dyval}, and facilitating LLMs in understanding graph data and graph reasoning may be a crucial step toward advancing their general capabilities in mathematics and code intelligence \cite{graphsilo, GraphPile}.

% \begin{table}[ht]
% \begin{tabularx}{7.8cm}{X}
% \toprule[1.2pt]
% \multicolumn{1}{c}{\textit{\textbf{Predecessor}}}\\ \hline
% \textbf{Input:} 
% Given a directed graph:\\
% Node 1 is connected to nodes 4, 9, 7, 10.\\
% Node 4 is connected to node 0.\\
% Node 9 is connected to nodes 6, 7.\\
% Node 7 is connected to nodes 1, 11, 2, 6.\\
% Node 10 is connected to nodes 1, 0, 11, 2, 6.\\
% Node 0 is connected to node 4.\\
% Node 5 is connected to nodes 11, 9, 7, 10.\\
% Node 2 is connected to node 0.\\
% Node 6 is connected to nodes 7, 0, 5, 4, 8.\\
% Node 3 is connected to nodes 7, 1, 11, 2.\\
% Node 8 is connected to nodes 1, 0, 4, 3.\\
% Which are the predecessor nodes of node 3? A predecessor of n is a node m such that there exists a directed edge from m to n.\\
% Let us answer this question step by step to make it correct. \\ \\ 
% \textbf{ChatGPT:} 
% % In the given directed graph, the predecessor nodes of Node 0 are the nodes from which there is a directed edge leading to Node 0. 
% Now, let's find the predecessors of node 3: Node 1 has a directed edge to node 3. \\
% Node 11 has a directed edge to node 3. \\
% Node 2 has a directed edge to node 3. \\
% Therefore, the predecessor nodes of node 3 are \colorbox{pink}{nodes 1, 11, and 2.} \XSolidBold \\ 
% \textbf{\modelname{}:} Let's solve it step by step.
% Nodes [2, 8, 0] connect to node 3, so the predecessor nodes of node 3 are
%  \colorbox{green}{[2, 8, 0].} \CheckmarkBold\\
% \bottomrule[1.2pt]
% \end{tabularx}
% \caption{Case studies on graph reasoning tasks that general LLMs fail to handle.}
% \label{tab:case_study}
% \end{table}

\begin{figure}[t]
    \centering
    \includegraphics[width = 0.46\textwidth]{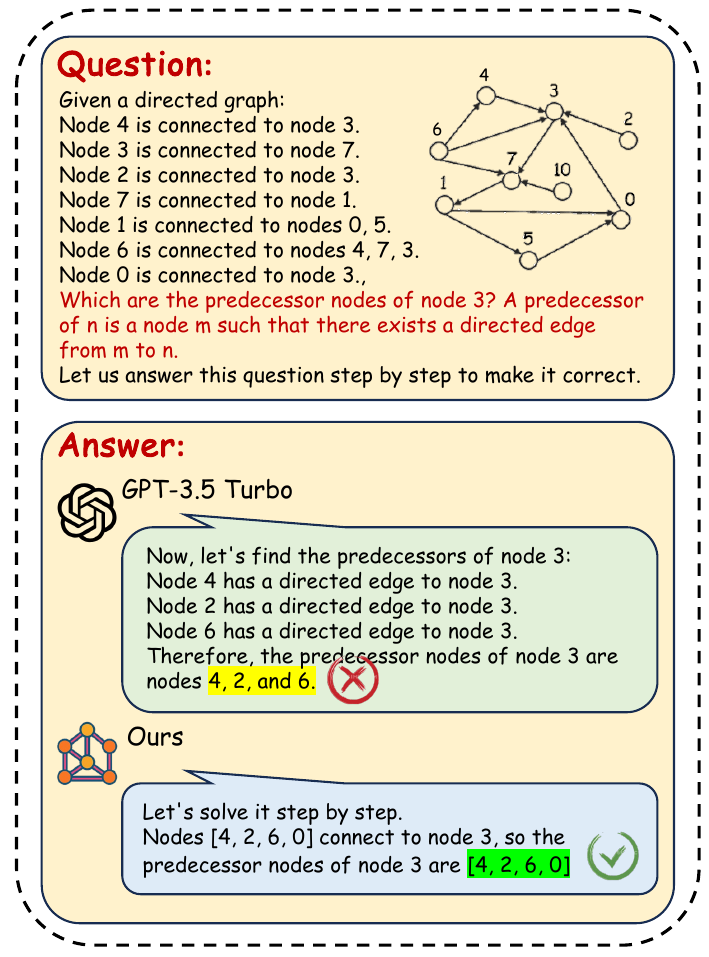}
    \caption{Case study on graph reasoning task that general LLMs like GPT-3.5 Turbo fail to handle}
    \vspace{-0.2cm}
    \label{fig:intro}
\end{figure}

In fact, numerous researchers have already started investigating the understanding capabilities of LLMs on graph data, such as GPT4Graph \cite{DBLP:journals/corr/abs-2305-15066}, NLGraph \cite{wang2023can}, GraphArena \cite{grapharena}, and others. Fatemi et al. \cite{fatemi2023talk} also examine the impact of different graph description languages on the understanding of graph data by LLMs. These exploratory works consistently find that, in most cases, LLMs perform significantly below expectations. As shown in Figure~\ref{fig:intro}, we conduct case studies on a classical graph reasoning task, implying that existing LLMs struggle to comprehend and address reasoning tasks based on graph data. In response to these challenges, a number of work begins to focus on enhancing the capabilities of LLMs on graph data. GraphLLM \cite{chai2023graphllm} and GraphGPT \cite{tang2023graphgpt} contemplate ways to enhance the performance of LLMs in graph-related tasks by concatenating encoded information from \textit{Graph Neural Networks} (GNNs) into LLMs. GraphWiz \cite{graphwiz} utilizes supervised fine-tuning (SFT) and direct preference optimization (DPO) to align LLMs with graph data. These works aid in understanding graph structure information but has only been tested on a limited set of tasks, with insufficient generalization capabilities. In summary, addressing the challenge of enabling LLMs to truly understand graph data remains an unresolved issue.

% Indeed, numerous researchers have begun investigating the understanding capabilities of large language models (LLMs) on graph data, such as GPT4Graph \cite{DBLP:journals/corr/abs-2305-15066}, NLGraph \cite{wang2023can}, Yamada et al. \cite{yamada2023evaluating}, exploring their performance on graph-based tasks. However, their findings reveal that, in most cases, the performance falls far below expectations. Fatemi et al. explored the impact of different graph description languages on LLMs' understanding of graph data. Despite these efforts, much of the existing work remains exploratory, lacking proposals for further enhancing LLMs' understanding of graphs. GraphLLM \cite{chai2023graphllm} and GraphGPT \cite{tang2023graphgpt} take a step further by incorporating encoding information from graph neural networks into LLMs to aid in comprehending graph structures. Nevertheless, their testing has been limited to a few tasks, revealing insufficient generalization capabilities. In summary, addressing the challenge of enabling LLMs to genuinely possess graph understanding and reasoning abilities, facilitating solutions for real-world graph-related tasks, remains a pressing and unresolved issue.

In this paper, to facilitate LLMs in comprehending graph-structured data and thereby equipping them with the ability to address graph-related tasks, we have curated and introduced a comprehensive dynamic benchmark – \textbf{{\benchmarkname}} designed for assessing and fine-tuning LLMs. Specifically, \benchmarkname{} comprises 21 classical tasks on graph structures with varying levels of difficulty, including elementary ones like node degree, connectivity, and common neighbors, as well as more complex ones such as maximum flow, shortest path, minimum
spanning tree, and so forth. 
Similar to the concept of \textit{Chain-of-Thought} (CoT) \cite{DBLP:conf/nips/KojimaGRMI22,DBLP:conf/nips/Wei0SBIXCLZ22}, to aid LLMs in better understanding the problem-solving approaches, we provide detailed reasoning steps for each question in the benchmark for training purposes. As a dynamic dataset, \benchmarkname{} provides diverse data generation pipelines and addresses the risk of data contamination. We conduct instruction tuning on our proposed benchmark \benchmarkname{} and construct \modelname{} with superior graph understanding capabilities. Experimental results demonstrate a significant improvement in \modelname{}'s performance across multiple classic graph reasoning tasks, which is comparable to GPT-3.5 Turbo. Building on this, we further turn to enhance the graph reasoning capability of LLMs based on \benchmarkname. Unlike graph understanding capability, graph reasoning capability requires ordered, multi-step logical inference that composes intermediate operations until the final answer is derived. With this goal in mind, we further introduce a label mask training strategy, encouraging LLMs to learn crucial node ID information during fine-tuning, and propose \modelnamestep{}, which demonstrates strong graph reasoning capability. 
% To assess whether LLMs genuinely possess universal graph understanding capabilities, we have also curated and released x graph data-related tasks xxxx based on real-world scenarios. Unlike xxx, xxxx is independent of training data and is entirely derived from real-world needs, simulating practical requirements in the real world. Through this evaluation approach, we aim to examine whether LLMs possess graph understanding capabilities and can apply them to solve real-world problems.

The contributions of this paper can be summarized as follows:
\begin{itemize}
    \item We propose a benchmark \benchmarkname{} for enhancing and evaluating the capabilities of LLM towards graph data. Different from existing benchmarks like NLGraph, GraphLLM, and others, our benchmark encompasses the most comprehensive set of tasks with diverse graph generation pipelines. We also provide a detailed chain of thought for each task, laying the groundwork for future advancement in enhancing graph reasoning capability.
    \item We propose an LLM namely \modelname{}, which is specifically for graph reasoning tasks through instruction tuning. Comprehensive experiments show that \modelname{} exhibits superior graph understanding abilities.
    \item Thanks to the detailed intermediate reasoning steps provided by \benchmarkname{}, we further propose a label mask training strategy and construct a model namely \modelnamestep{}. Extensive experiments verify the effectiveness of our strategy and the prominent graph reasoning capability of \modelnamestep{}. 
\end{itemize}

%% file: related_work.tex
\section{Related Work}
\subsection{LLMs on Graph Reasoning}

With the rise of LLMs, researchers are increasingly exploring the capability of LLMs on solving graph reasoning tasks like DFS and the shortest path. Existing research can be broadly categorized into three domains: 1) \textbf{Empirical performance evaluation}. Early researchers conduct empirical performance evaluations, such as GPT4Graph \cite{DBLP:journals/corr/abs-2305-15066}, NLGraph \cite{wang2023can}, GraphArena \cite{grapharena}, GraphQA \cite{fatemi2023talk}, and GraphOmni \cite{graphomni}. These early works pave the way for applying LLMs to the field of graph data and consistently reveal the current inadequacy of LLMs in understanding graph data. 2) \textbf{Fine-tuning for graph-specific reasoning.} Many researchers have started fine-tuning LLMs on graph data to enhance their reasoning capability on graph-related tasks, as seen in works like GraphLLM \cite{chai2023graphllm}, GraphWiz \cite{graphwiz}, and GraphSilo \cite{graphsilo} etc. However, their scope has remained relatively limited to several specific tasks, and the challenge of enabling LLMs to possess a universal understanding of graph data remains an urgent problem to be addressed. 3) \textbf{Tool augmentation and code generation}. Different from the existing work, the third line of researchers do not ask LLMs to solve graph reasoning tasks in an explicit step-by-step reasoning manner, but utilizes tool augmentation (e.g., code generation) to equip LLMs with explicit graph algorithmic skills \cite{GCoder, Graphforge}. Our work, \benchmarkname{}, belongs to the second category. Our aim is to help LLMs genuinely master graph reasoning by advancing fine-tuning methodologies and promoting universal understanding of graph data.

\begin{figure*}[t]
    \centering
    \begin{minipage}[t]{0.95\textwidth}
    \includegraphics[width = \textwidth]{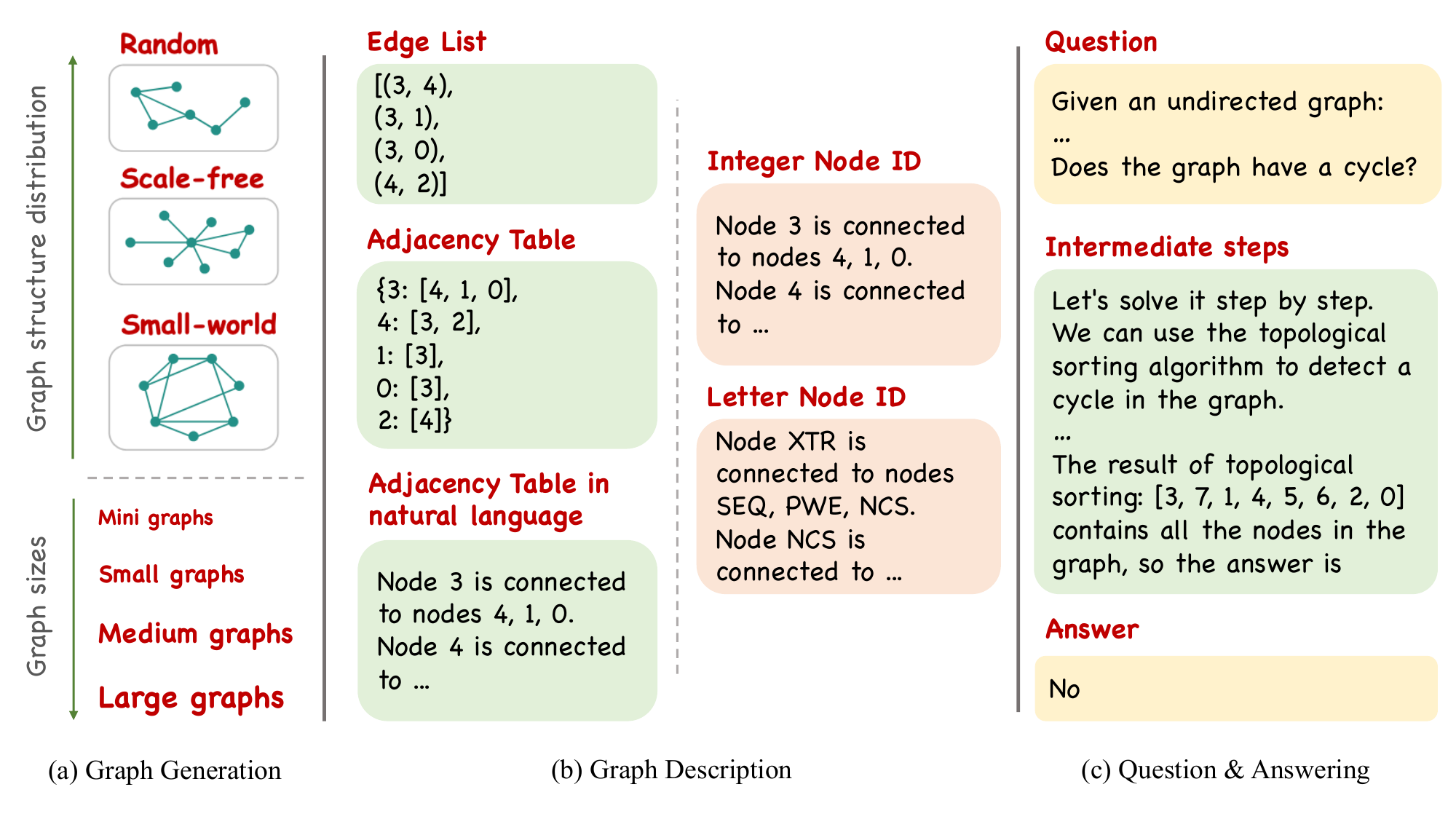}
    \end{minipage}%
    \vspace{-0.4cm}
    \caption{The overview of \benchmarkname{} benchmark. We provide diverse options during the construction of \benchmarkname{}, including graph structure distributions, graph sizes, graph description languages, and node IDs. For improving the reasoning capability of LLMs, \benchmarkname{} also provides precise intermediate results for each task.}
    \label{fig:pipeline}
\end{figure*}

\subsection{LLM Instruction Tuning}

Although LLMs are widely applied in various fields such as conversation \cite{CPsyCoun}, mathematics \cite{wizardlm}, and code programming \cite{bigcodebench} due to their outstanding generalization abilities, researchers have found that fine-tuning with specific instructions can further enhance LLMs' performance in more targeted domains. For instance, Bao et al. \cite{bao2023tallrec} utilize instruction tuning to apply LLMs to recommendation scenarios. Li et al. \cite{li2023ecomgpt} create a dedicated dataset, EcomInstruct, for the e-commerce domain and fine-tune a specialized LLM, EcomGPT. Moreover, researchers have explored the application of LLMs in various areas such as math \cite{t1,wizardmath}, healthcare \cite{HealthLLM,mmedc}, law \cite{yue2024lawllm}, and code \cite{DBLP:journals/corr/abs-2306-08568}. These successful attempts demonstrate that fine-tuning LLMs can enhance their understanding of different types of data and improve their instruction-following capability, enabling their application in diverse domains.

%% file: benchmark.tex
\section{The {\benchmarkname} Benchmark}
\label{sec:benchmark}

\begin{figure}[t]
    \centering
    \begin{minipage}[t]{0.35\textwidth}
    \includegraphics[width = 0.98\textwidth]{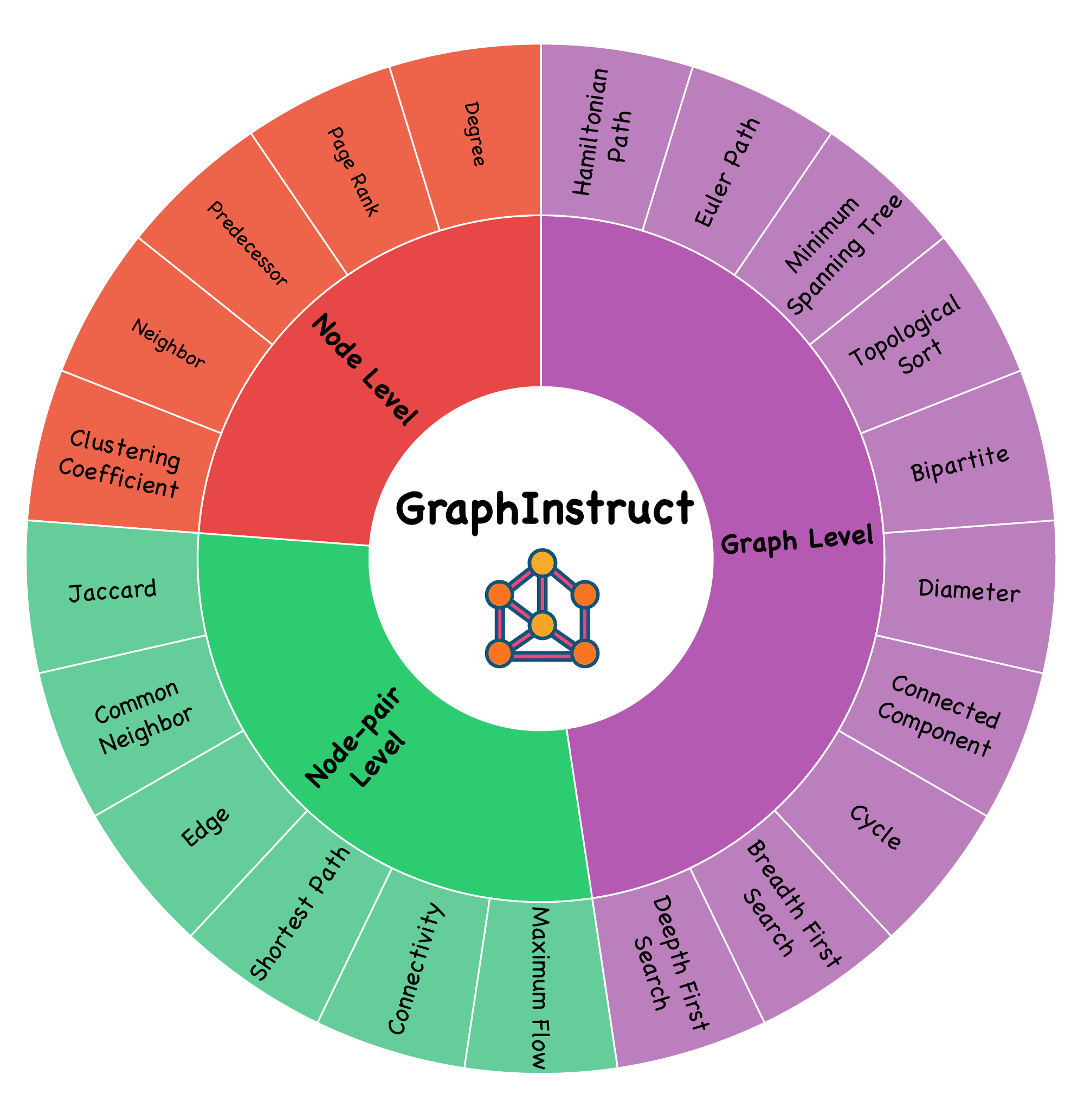}
    \end{minipage}%
    \caption{The complete task schema of \benchmarkname{}}
    \vspace{-0.2cm}
    \label{fig:schema}
\end{figure}

In this section, we will introduce our \benchmarkname{} benchmark, which is designed to evaluate and enhance the graph understanding and graph reasoning capability of LLMs. In general, \benchmarkname{} covers various graph structure distributions, graph sizes, and graph descriptions. The overview of \benchmarkname{} is illustrated in Figure~\ref{fig:pipeline}, where samples are dynamically generated with various options. 
As shown in Figure~\ref{fig:schema}, we comprehensively consider 21 classic graph reasoning tasks in \benchmarkname{}, encompassing tasks at \textit{Node level}, \textit{Node-pair level}, and \textit{Graph level}, respectively, and cover most classical graph reasoning tasks such as \textit{Depth-First Search} (DFS) and Shortest Path. The following subsections will further introduce the details of \benchmarkname{}.

\subsection{Graph Generation}
As the foundation in \benchmarkname{}, the graphs generated in \benchmarkname{} should be as diverse as possible, so that the performance on these graphs could reflect the intrinsic capability of LLMs. 
To this end, we provide various kinds of options during the graph generation process, as shown in Figure \ref{fig:pipeline} (a), and can be summarized as follows:
% the graphs in \benchmarkname{} are generated with multiple protocols as follows.

\paragraph{Graph Structure Distribution} To provide a diverse range of graph structures, we mainly consider three representative graph structure distributions, including (1) Erdős–Rényi random networks; (2) Barabási–Albert scale-free networks \cite{doi:10.1126/science.286.5439.509}; and (3) small-world networks \cite{watts1998collective}: 

\begin{itemize}[leftmargin=1em]
    \item \textbf{Erdős–Rényi random networks}: Given $N$ nodes, they are connected pairwise with a probability of $p$ to form a graph.
    \item \textbf{Barabási–Albert scale-free networks}: Networks in which the node degree distribution follows a power-law distribution. Various networks such as social networks, biological networks, and trade networks, exhibit scale-free network characteristics.
    \item \textbf{Small-world networks}: In small-world networks, most nodes are not directly connected to each other, but the majority of nodes can be reached from any other node through a few intermediate steps.
\end{itemize}

In addition, both directed and undirected graphs are considered and randomly generated in \benchmarkname{} during graph generation, providing diverse graph structure distributions.

\paragraph{Graph Size} Following the line of previous work \cite{chai2023graphllm,wang2023can}, \benchmarkname{} consists of four levels of graph size, namely mini graphs, small graphs, medium graphs and large graphs, consisting of 5-7 nodes, 8-15 nodes, 16-25 nodes and 26-35 nodes, respectively. 
Intuitively, tasks with larger graph sizes are always incorporated with higher difficulty levels because of the increase in context length and required reasoning steps. The impact of graph size will be further analyzed in our experiments.

\subsection{Graph Description}

\benchmarkname{} is a text-based dataset that transforms the generated graphs into diverse \textit{Graph Description Languages} (GDL). Given the prior investigation provided by \cite{fatemi2023talk}, {\benchmarkname} provides three effective graph description languages as shown in Figure \ref{fig:pipeline} (b): edge list, adjacency table, and adjacency table in natural language (the default setting during training). 
% When the graph is undirected, we only record edge $(u, v)$ once. 
Besides different graph description languages, we also provide two optional node ID representations, including an integer ID like '1' and random letters like 'XRT'. 

\subsection{Intermediate Steps Generation}

To help the LLMs better accomplish the graph reasoning tasks in a CoT manner, we generate detailed intermediate reasoning steps for each training sample through graph algorithms. 
As illustrated in Figure \ref{fig:pipeline} (c), the stepwise information encodes accurate intermediate steps, which can be used as either supervision signals or CoT prompts. With the help of intermediate steps, we can empower LLMs with stronger graph reasoning capability, which will be discussed in Section~\ref{sec:step}.

\subsection{Task Definition}
For each sample, \benchmarkname{} $\mathcal{D}$ provides four elements, including the graph $\mathcal{G}$, query $\mathcal{Q}$, answer $\mathcal{A}$ and intermediate reasoning steps $\mathcal{S}$. For each randomly generated graph $\mathcal{G} = (\mathcal{V}, \mathcal{E})$, $\mathcal{V}$ is the set of nodes and $\mathcal{E}$ is the set of edges. In the following, we briefly introduce the definitions for the 21 graph reasoning tasks involved in \benchmarkname{}:  

\begin{itemize}[leftmargin=1em]
\setlength{\itemsep}{0pt}
\setlength{\parsep}{0pt}
\setlength{\parskip}{0pt}
\item \textbf{Neighbor} Given a node $u \in \mathcal{V}$ , output its neighbor nodes $\mathcal{N}(u)$. 

\item \textbf{Degree} Given a node $u \in \mathcal{V}$, output the degree of it: $|\mathcal{N}(u)|$.

\item \textbf{Predecessor} Given a directed graph $\mathcal{G}$ and a node $u \in \mathcal{V}$, output the set of predecessor nodes of $u$: $\mathcal{P}(u) = \{ v | (v, u) \in \mathcal{E} \}$.

\item \textbf{PageRank} Find the node with the largest PageRank value \cite{DBLP:conf/www/JehW03} after several rounds of iteration. The damping value is 0.85, and the iteration number is 3. The initialized PageRank value for each node is $1/|\mathcal{V}|$.

\item \textbf{Clustering Coefficient} Calculate the clustering coefficient value of node $u \in \mathcal{V}$. For a node $u$, let $T$ be the edge number among its neighbors and $D$ its degree (out-degree in directed graphs). The clustering coefficient of $u$ is $C=\frac{T}{D(D-1)}$ for directed graphs and $C=\frac{2T}{D(D-1)}$ for undirected graphs.

\item \textbf{Common Neighbor} Given a node pair ($u$, $v$), output the number of common neighbors: $CN = |\mathcal{N}(u) \cap \mathcal{N}(v)|$. 

\item \textbf{Jaccard} Given a node pair ($u$, $v$), compute the Jaccard Coefficient $J$ , which can be formulated as $J(u, v) = \frac{|\mathcal{N}(u) \cap \mathcal{N}(v)|}{|\mathcal{N}(u) \cup \mathcal{N}(v)|}$. 

\item \textbf{Edge} Determine if $<u, v>$ exists in the edge set $\mathcal{E}$.

\item \textbf{Shortest Path} Given a weighted graph $\mathcal{G}$ (we randomly set an integer weight value from 1 to 10 for each edge), output the distance of the shortest path from node $u$ to node $v$. 

\item \textbf{Connectivity} Determine if there exists a sequence of adjacent nodes leading from node $u$ to node $v$. 

\item \textbf{Maximum Flow} Given a weighted graph $\mathcal{G}$, output the maximum amount of flow from node $u$ to node $v$. 

\item \textbf{Depth-First Search} Perform a depth-first traversal of the graph starting from node $u$, and return the list of nodes in the order they are visited.

\item \textbf{Breath-First Search} Perform a breadth-first traversal of the graph starting from node $u$, and return the list of nodes in the order they are explored.

\item \textbf{Cycle} Determine if the graph $\mathcal{G}$ contains a cycle, which is defined as a closed path that starts and ends at the same node, with all intermediate nodes and edges visited exactly once. 

\item \textbf{Connected Component} Given a node $u \in \mathcal{V}$, output the set of nodes in the same connected component of $u$. 

\item \textbf{Diameter} Calculate the diameter of the graph, which is the maximum distance over all pairs of nodes in the graph. 

\item \textbf{Bipartite} Given a bipartite graph $\mathcal{G}$, find a maximum matching (a list of node pairs). 

\item \textbf{Topological Sort} Given a directed acyclic graph $\mathcal{G}$, produce a topological ordering of its nodes—i.e., a linear sequence in which each node appears before all nodes to which it has outgoing edges.

\item \textbf{Minimum Spanning Tree (MST)} Given a graph $\mathcal{G}$, compute the total weight of its minimum spanning tree, which connects all nodes with the minimal possible sum of edge weights. 

\item \textbf{Euler Path} Find an Euler path in the graph $\mathcal{G}$. An Euler path in a graph is a path that traverses each edge exactly once, but not necessarily every node. The starting node and ending node may be different. 

\item \textbf{Hamiltonian Path} Find a Hamiltonian path in the graph $\mathcal{G}$. A Hamiltonian path in a graph is a path that visits each node exactly once, traversing along edges, and the starting node and ending node may be different.

\end{itemize}

% \begin{figure}[ht]
%     \centering
%     \includegraphics[width = 0.44\textwidth]{figure/steps.pdf}
%     \caption{Intermediate Results for Training}
%     \label{fig:steps}
% \end{figure}

\subsection{Evaluation}

We parse the LLM's output and use exact-match accuracy for evaluation. 
There are mainly the following kinds of output type: boolean (e.g. Connectivity), integer (e.g. Degree), float (e.g. Jaccard), single node (e.g. PageRank), node list (e.g. Depth-First Search), node set (e.g. Neighbor), and edge list (e.g. Bipartite).
% We require the output of LLM to be a valid answer. 
For float type output, we require a relative error of 3\% or less. 
For tasks where multiple solutions exist (e.g. Depth-First Search), we examine if the output is a valid answer. 

\begin{table}[t]
\centering
\caption{Comparisons of \benchmarkname{} among other related datasets}
\label{tab:dataset}
\resizebox{8.5cm}{!}{%
\begin{tabular}{l|ccccc|c}
\toprule[1.2pt]
\multirow{2}{*}{} & \multirow{2}{*}{\textbf{Train Corpus}} & \multirow{2}{*}{\textbf{CoT Steps}} 
   & \multicolumn{3}{c|}{\textbf{Sample Diversity}} & \multirow{2}{*}{\textbf{Tasks}} \\ \cline{4-6}
   & & & \textbf{Graph Structure} & \textbf{Node ID} & \textbf{Graph Description} & \\
\hline
NLGraph \cite{wang2023can} & \textcolor{red}{\ding{55}} & \textcolor{red}{\ding{55}} & \textcolor{red}{\ding{55}} & \textcolor{red}{\ding{55}} & \textcolor{green}{\ding{51}} & 8 \\ 
GraphQA \cite{fatemi2023talk} & \textcolor{red}{\ding{55}} & \textcolor{red}{\ding{55}} & \textcolor{red}{\ding{55}} & \textcolor{green}{\ding{51}} & \textcolor{green}{\ding{51}} & 12 \\ 
GraphOmni \cite{graphomni} & \textcolor{red}{\ding{55}} & \textcolor{red}{\ding{55}} & \textcolor{green}{\ding{51}} & \textcolor{red}{\ding{55}} & \textcolor{green}{\ding{51}} & 6 \\
GraphArena \cite{grapharena} & \textcolor{red}{\ding{55}} & \textcolor{red}{\ding{55}} & \textcolor{green}{\ding{51}} & \textcolor{green}{\ding{51}} & \textcolor{green}{\ding{51}} & 10 \\
NLGift \cite{nlgift} & \textcolor{green}{\ding{51}} & \textcolor{red}{\ding{55}} & \textcolor{green}{\ding{51}} & \textcolor{green}{\ding{51}} & \textcolor{green}{\ding{51}} & 4 \\
GraphWiz \cite{graphwiz} & \textcolor{green}{\ding{51}} & \textcolor{green}{\ding{51}} & \textcolor{red}{\ding{55}} & \textcolor{red}{\ding{55}} & \textcolor{red}{\ding{55}} &  9 \\
GraphSilo \cite{graphsilo} & \textcolor{green}{\ding{51}} & \textcolor{green}{\ding{51}} & \textcolor{green}{\ding{51}} & \textcolor{red}{\ding{55}} & \textcolor{red}{\ding{55}} & 13 \\
\hline
\rowcolor[RGB]{234, 238, 234}
\textbf{\benchmarkname{} (Ours)} & \textcolor{green}{\ding{51}} & \textcolor{green}{\ding{51}} & \textcolor{green}{\ding{51}} & \textcolor{green}{\ding{51}} & \textcolor{green}{\ding{51}} & \textbf{21} \\
\bottomrule[1.2pt]
\end{tabular}}
\vspace{-0.2cm}
\end{table}

\subsection{\benchmarkname{} Statistics}

As summarized in Table~\ref{tab:dataset}, we compare \benchmarkname{} with several existing datasets on graph reasoning tasks. To the best of our knowledge, \benchmarkname{} is the most comprehensive dataset, encompassing 21 classical tasks and providing detailed intermediate reasoning steps for each sample. In addition, \benchmarkname{} could provide a variety sample distribution due to the diverse options during graph generations, making it a fertile soil for developing LLMs' graph understanding capability.

%% file: model.tex
\begin{figure*}[t]
    \centering
    \begin{minipage}[t]{\textwidth}
    \includegraphics[width = \textwidth]{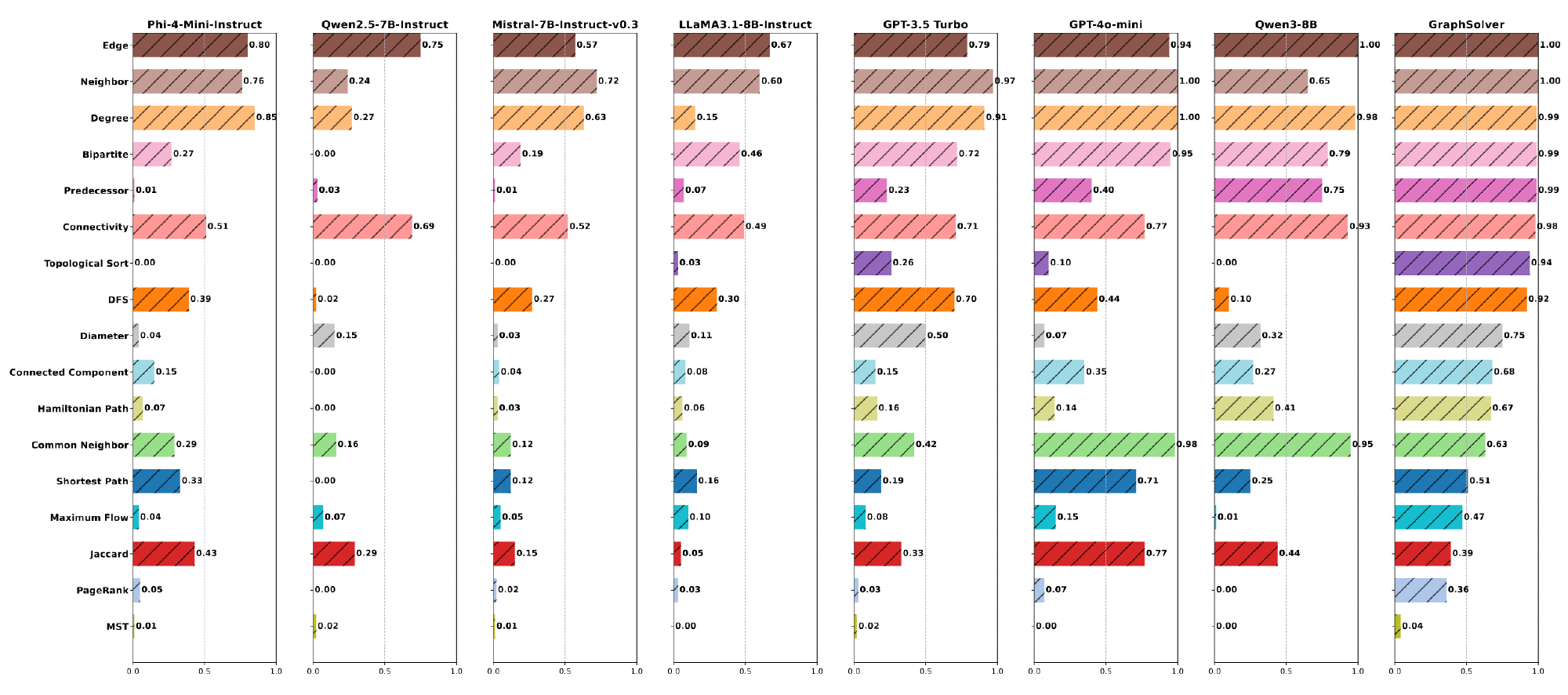}
    \end{minipage}%
    \caption{Accuracy comparison of several LLMs on 17 in-domain tasks of \benchmarkname{}}
    \label{fig:main_experiment}
\end{figure*} 

\section{\modelname{}: Towards Graph Understanding}
\label{sec:train}

Thanks to the diversity inherent in \benchmarkname{}, we posit that this benchmark holds the potential to enhance LLMs with graph understanding capability. With LLaMA-3.1-8B-Instruct as the foundational model, we first construct a model named \modelname{} through instruction tuning on the \benchmarkname{}. In detail, given a sample $<\mathcal{G}, \mathcal{Q}, \mathcal{A}> \in \mathcal{D}$, the \modelname{} with parameters $\theta$ is trained to minimize the following loss function:
\begin{equation}
    \mathcal{L}=-\mathbb{E}_{<\mathcal{G},\mathcal{Q},\mathcal{A}>\sim \mathcal{D}}{\rm log}P(\mathcal{A}|(\mathcal{G}, \mathcal{Q});\theta).
\end{equation}
Note that, at this stage, our main aim is to develop fundamental graph understanding capability for LLMs, thus we do not incorporate reasoning steps $\mathcal{S}$ here. The role of intermediate reasoning steps will be analyzed in Section~\ref{sec:step}. 

Specifically, we employ LoRA \cite{DBLP:conf/iclr/HuSWALWWC22} as the fine-tuning strategy, which is adapted to both attention modules and feed-forward networks, and the rank dim is set to 8. Adam \cite{DBLP:journals/corr/KingmaB14} is used as the optimizer, with the learning rate set of $1e^{-4}$ and weight decay of 0. The batch size is set to 8 during the training process, and a cosine learning rate scheduler is utilized. We fine-tune the model for 4 epochs based on LLaMAFactory \cite{llamafactory}, and all experiments are implemented on a machine with 4 32GB NVIDIA V100 GPUs. As an 8B-parameter LLaMA model, the computation cost of \modelname{} is actually comparable to other LLMs of similar size. To further facilitate the inference efficiency, we employ vLLM \cite{vllm} techniques for all models, which greatly reduces the inference time cost.

%% file: experiment.tex
\subsection{Experimental Settings}

\paragraph{Dataset split} The tasks in \benchmarkname{} are divided into two groups according to whether the tasks appear in the training data during fine-tuning the model, which we call \textit{In-domain tasks} and \textit{Out-of-domain tasks}. In detail, we choose four relatively difficult tasks named \textit{BFS}, \textit{Cycle}, \textit{Clustering Coefficient}, and \textit{Euler Path} as the Out-of-domain tasks, and the rest as In-domain tasks. The reasons are two-fold: 1) The four tasks are inherently challenging, demanding advanced graph comprehension and reasoning capabilities from LLMs for successful completion; 2) These four tasks comprehensively cover node-level, node pair-level, and graph-level aspects, and the required output format for each task involves boolean, numerical values, and node sets, which are quite diverse. For each in-domain task, 800 samples are randomly selected as the training set, which consists of 400 samples with tiny-level graph size and 400 samples with small-level graph size. As for test, we randomly select 100 samples as the test set for each task, which results in a final test set with 2100 samples. Note that, graphs generated in \benchmarkname{} are described with adjacency tables in natural language and integer node ID by default. The generalization analysis to more graph description options will be given in Section~\ref{sec:generalization}.

\paragraph{Baselines} Besides LLaMA-3.1-8B-Instruct\footnote{https://huggingface.co/meta-llama/Llama-3.1-8B-Instruct}, we also evaluate multiple powerful LLMs with comparable model scale, including Qwen-3-8B\footnote{https://huggingface.co/Qwen/Qwen3-8B}, Qwen-2.5-7B-Instruct\footnote{https://huggingface.co/Qwen/Qwen2.5-7B-Instruct}, Phi-4-mini-Instruct\footnote{https://huggingface.co/microsoft/Phi-4-mini-instruct}, Mistral-7B-Instruct-v0.3\footnote{https://huggingface.co/mistralai/Mistral-7B-Instruct-v0.3}, and two proprietary LLMs namely GPT-3.5 Turbo and GPT-4o-mini\footnote{https://openai.com/index/gpt-4o-system-card/}. For all baselines, the temperature $\tau$ is set to 0 for obtaining determined answers. It is worth noting that we do not include GNNs like GCN \cite{gcn} and GraphSAGE \cite{graphsage} here due to their fundamentally different paradigms \cite{grapharena} and the difference of downstream tasks. GNNs are task-specific models and are based on network data, while we aim to introduce a model that has a general graph understanding capability and can handle different graph inputs. In this way, we believe that our model is complementary rather than competitive to traditional GNNs.

\subsection{Main Experiments}

In this part, we first investigate the performance of \modelname{} and several powerful LLMs on \benchmarkname{}. Specifically, the zero-shot accuracy on 17 in-domain tasks is reported, since we contend that the zero-shot performance can better reflect the LLM's fundamental comprehension of graph data. The results are shown in Figure~\ref{fig:main_experiment}, from which we can observe that:
\begin{itemize}[leftmargin=1em]
    \item For open-sourced LLMs like Phi-4-mini-Instruct and Qwen-2.5-7B-Instruct with relatively small model sizes, their performance on \benchmarkname{} is poor, which implies LLMs' lack of capability towards graph data, especially for models with fewer parameters.
    \item Following fine-tuning, \modelname{} demonstrates a substantial improvement over LLaMA-3.1-8B-Instruct across the majority of tasks, showcasing performance that is comparable or even superior to GPT-3.5 Turbo. For example, LLaMA-3.1-8B-Instruct achieves a mere 15\% accuracy in the \textit{Neighbor} task, whereas \modelname{} attains an impressive 99\% accuracy, surpassing even the performance of GPT-3.5 Turbo and being comparable with GPT-4o-mini.
    \item Despite the impressive performance on certain tasks, such as straightforward graph reasoning tasks like \textit{Degree} and \textit{Neighbors}, both GPT-4o-mini and \modelname{} exhibit less satisfactory results when tackling more complex graph reasoning tasks, including \textit{MST} and \textit{PageRank}. This underscores the existing challenges that LLMs face in comprehending graph-structured data. More efforts are still in demand in the future.
\end{itemize}

\begin{figure}[t]
    \centering
    \begin{minipage}[t]{0.43\textwidth}
    \includegraphics[width = \textwidth]{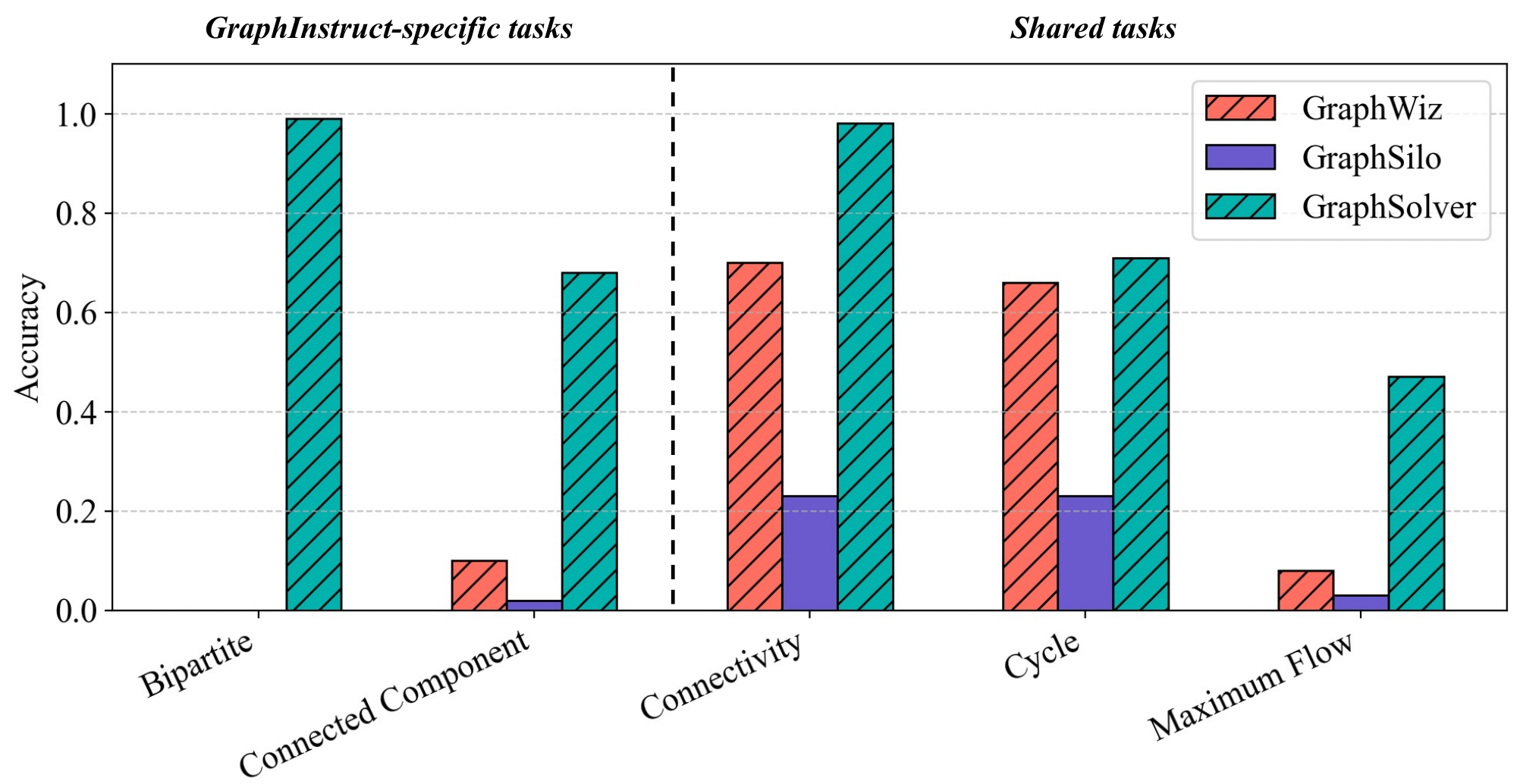}
    \end{minipage}%
    \caption{Performance comparisons on both \benchmarkname{}-specific tasks and shared tasks with two competitor models}
    \label{fig:comparison}
\end{figure}

\subsection{Comparison with Other Competitors}
In this section, we aim to compare \modelname{} with two highly related competitors, namely GraphWiz \cite{graphwiz} and GraphSilo \cite{graphsilo}. Specifically, we conduct evaluations on (1) tasks that are shared across all works — \textit{Cycle}, \textit{Connectivity}, and \textit{Maximum Flow}, and (2) two tasks that are exclusive to the \benchmarkname{} —\textit{Bipartite} and \textit{Connected Component}. The results are shown in Figure~\ref{fig:comparison}, where \modelname{} consistently achieves the best performance on all tasks, especially on two tasks that only \benchmarkname{} includes. Such observations highlight the significance of introducing a comprehensive training corpus for developing the capability on graph reasoning tasks for LLMs.

\subsection{Results with Prompt Engineering}

\begin{table}[t]
\caption{The effect of different prompt engineering techniques}
\resizebox{8.5cm}{!}{
\begin{tabular}{clccccc}
\toprule[1.2pt]
\multicolumn{2}{l}{}                                       & \textbf{DFS} & \textbf{Degree} & \textbf{Common Neighbor} & \textbf{Connectivity} & \textbf{Predecessor} \\ \hline
\multicolumn{2}{c}{\textbf{GPT-3.5 Turbo}}                 & 0.37         & 0.98            & 0.77                     & 0.71                  & 0.19                 \\ \hline
\multirow{3}{*}{\textbf{GraphSolver}} & \textbf{Zero-shot} & 0.92         & 0.99            & 0.63                     & 0.98                  & 0.99                 \\
                                      & \textbf{One-shot}  & 0.91         & 0.99            & 0.64                     & 0.97                  & 0.97                 \\
                                      & \textbf{CoT}       & 0.90         & 0.99            & 0.64                     & 0.98                  & 0.95                 \\ \bottomrule[1.2pt]
\end{tabular}}
\label{tab:prompt}
\end{table}

Table~\ref{tab:prompt} presents the effect of prompt engineering techniques such as Few-shot and CoT \cite{DBLP:conf/nips/KojimaGRMI22,DBLP:conf/nips/Wei0SBIXCLZ22} across five graph reasoning tasks: \textit{DFS}, \textit{Degree}, \textit{Common Neighbor}, \textit{Connectivity}, and \textit{Predecessor}. For comparison, we also report the zero-shot results of GPT-3.5 Turbo as a reference. It can be seen that all three prompting-enhanced variants of \modelname{} — zero-shot, one-shot, and zero-shot CoT —exhibit substantially stronger performance across most tasks. However, the differences among the three prompting strategies were minimal. Zero-shot, one-shot, and CoT yielded nearly identical results across all metrics, with only marginal fluctuations. We assume that the limited impact of prompt engineering might be attributed to the intrinsic reasoning capabilities already embedded in the fine-tuned model - \modelname{}.

\subsection{Generalization Analysis}
\label{sec:generalization}
\begin{figure}[t]
    \centering
    \begin{minipage}[t]{0.495\textwidth}
    \includegraphics[width = \textwidth]{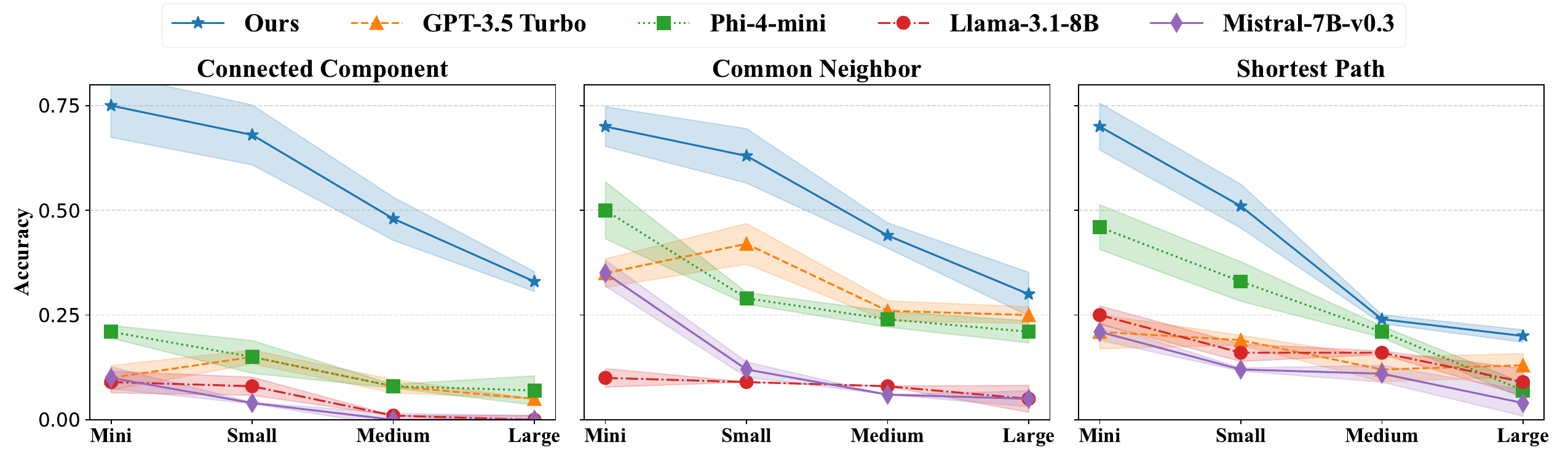}
    \end{minipage}%
    \caption{Zero-shot performance on 3 classical graph reasoning tasks in \benchmarkname{} with different graph sizes}
    \vspace{-1em}
    \label{fig:graph_size}
\end{figure}

\begin{table*}[ht]
\centering
\caption{Zero-shot performance (Accuracy) on 8 classical graph reasoning tasks in \benchmarkname{} with different graph description languages}
\label{tab:gml}
\small
\renewcommand{\arraystretch}{1.05}
\resizebox{17cm}{!}{%
\begin{tabular}{lcccccccccccc}
\toprule
\multirow{2}{*}{\textbf{Task}} & \multicolumn{2}{c}{\textbf{GPT-3.5 Turbo}} & \multicolumn{2}{c}{\textbf{Phi-4-mini}} & \multicolumn{2}{c}{\textbf{Qwen-2.5-7B}} & \multicolumn{2}{c}{\textbf{Mistral-7B-v0.3}} & \multicolumn{2}{c}{\textbf{LLaMA-3.1-8B}} & \multicolumn{2}{c}{\textbf{\modelname{}}} \\
\cmidrule(lr){2-3}\cmidrule(lr){4-5}\cmidrule(lr){6-7}\cmidrule(lr){8-9}\cmidrule(lr){10-11}\cmidrule(lr){12-13}
                              & \textbf{Adjacency} & \textbf{Edge List} & \textbf{Adjacency} & \textbf{Edge List} & \textbf{Adjacency} & \textbf{Edge List} & \textbf{Adjacency} & \textbf{Edge List} & \textbf{Adjacency} & \textbf{Edge List} & \textbf{Adjacency} & \textbf{Edge List} \\
\midrule
\textbf{Shortest Path}        & 0.06&  0.33&  0.42&  0.34&  0.01&  0.01&  0.13&  0.09&  0.17&  0.12&  0.41&  \textbf{0.43}  \\
\textbf{Hamiltonian Path}     & 0.16&  0.15&  0.11&  0.09&  0.00&  0.01&  0.05&  0.01&  0.01&  0.02&  \textbf{0.59}&  0.19  \\
\textbf{Edge}                 & 0.82&  0.76&  0.65&  0.69&  0.62&  0.73&  0.51&  0.37&  0.76&  0.76&  \textbf{1.00}&  0.81  \\
\textbf{Connected Component}  & 0.25&  0.16&  0.18&  0.12&  0.00&  0.00&  0.01&  0.00&  0.12&  0.10&  0.43&  \textbf{0.44}  \\
\textbf{Connectivity}         & 0.78&  0.74&  0.37&  0.58&  0.57&  0.59&  0.38&  0.10&  0.64&  0.68&  \textbf{0.92}&  0.87  \\
\textbf{Predecessor}          & 0.29&  0.35&  0.02&  0.00&  0.00&  0.02&  0.03&  0.04&  0.11&  0.10&  0.81&  \textbf{0.95}  \\
\textbf{Common Neighbor}      & 0.42&  0.31&  0.22&  0.18&  0.31&  0.49&  0.31&  0.24&  0.10&  0.03&  \textbf{0.62}&  0.39  \\
\textbf{Diameter}             & 0.45&  0.48&  0.10&  0.08&  0.12&  0.04&  0.26&  0.13&  0.04&  0.06&  \textbf{0.71}&  0.58  \\
\bottomrule
\end{tabular}%
}
\end{table*}

This section focuses on analyzing the generalization capabilities of \modelname{} from the following four perspectives: increasing graph sizes, distinct graph description languages, various node representation methods, and out-of-domain tasks.

\paragraph{Graph size} Throughout the training process, \modelname{} has exclusively encountered tiny and small graphs. Consequently, we aim to explore whether \modelname{} maintains its graph understanding capabilities when confronted with graphs featuring more nodes and intricate structures. In detail, we conduct experiments on graphs of different scales across three graph reasoning tasks, including \textit{Connected Component}, \textit{Common Neighbor}, and \textit{Shortest Path}.
It can be seen from Figure~\ref{fig:graph_size} that: 1) As the number of nodes and the complexity of the graph structure increase, all LLMs experience a decline in performance. This aligns with our expectations, given that more complex graphs inherently pose more challenging reasoning tasks for LLMs. 2) Notably, despite the increased difficulty, \modelname{} consistently outperforms other LLMs, even on previously unseen graph sizes such as \textit{medium} and \textit{large}. This observation underscores the robustness of \modelname{} in handling increasing graph sizes.

\paragraph{Graph description language} To delve deeper into the generalization capabilities of \modelname{} across diverse graph description languages, we conduct experiments on eight classic graph reasoning tasks using two unseen graph description languages: edge list and adjacency table (adj). The results, as illustrated in Table~\ref{tab:gml}, reveal that although there is a slight performance decrease when compared to that using the adjacency table in natural language, \modelname{} still consistently surpasses other LLMs, even including proprietary models like GPT-3.5 Turbo. The above observations verify \modelname{}'s capability to retain a substantial portion of its graph understanding capabilities when facing different graph description languages.

\begin{table}[t]
\centering
\caption{Zero-shot performance (Accuracy) on 8 classical graph reasoning tasks in \benchmarkname{} with different node ID representation}
\small
\resizebox{8.5cm}{!}{%
\begin{tabular}{lccccc c}
\toprule
\textbf{Task} & \textbf{GPT-3.5 Turbo} & \textbf{Phi-4-mini} & \textbf{Qwen-2.5-7B} & \textbf{Mistral-7B-v0.3} & \textbf{LLaMA-3.1-8B} & \textbf{\modelname{}} \\
\midrule
\textbf{Shortest Path}          & 0.33 & 0.03 & 0.01 & 0.24 & 0.17 & \textbf{0.54} \\
\textbf{Hamiltonian Path}& 0.16 & 0.04 & 0.00 & 0.00 & 0.05 & \textbf{0.31} \\
\textbf{Edge}            & 0.80 & 0.82 & 0.82 & 0.14 & 0.87 & \textbf{1.00} \\
\textbf{Connected Component}        & 0.18 & 0.21 & 0.00 & 0.04 & 0.11 & \textbf{0.52} \\
\textbf{Connectivity}    & 0.58 & 0.52 & 0.58 & 0.02 & 0.54 & \textbf{0.96} \\
\textbf{Predecessor}     & 0.19 & 0.02 & 0.04 & 0.03 & 0.10 & \textbf{0.79} \\
\textbf{Common Neighbor} & 0.37 & 0.46 & 0.16 & 0.19 & 0.15 & \textbf{0.61} \\
\textbf{Diameter}        & 0.34 & 0.09 & 0.02 & 0.13 & 0.07 & \textbf{0.62} \\
\bottomrule
\end{tabular}%
}
\label{tab:node_id}
\end{table}

\paragraph{Node ID representation} In contrast to the integer node IDs utilized in the training process (e.g., Node 1) of \modelname{}, we perform generalization tests by substituting node IDs in the test set with random letters (e.g., Node XTE). The results on eight representative graph reasoning tasks, depicted in Table~\ref{tab:node_id}, highlight that even with the altered node IDs, \modelname{} consistently outperforms the baseline models like Phi-4-mini-Instruct and GPT-3.5 Turbo, affirming its retention of graph understanding capabilities.

\begin{table}[t]
\centering
\caption{Zero-shot performance (Accuracy) on 4 OOD graph reasoning tasks in \benchmarkname{}}
\resizebox{7cm}{!}{
\begin{tabular}{lcccc}
\toprule
\textbf{Model} & \textbf{BFS} & \textbf{Clustering Coefficient} & \textbf{Cycle} & \textbf{Euler Path} \\
\midrule
\textbf{GPT-3.5 Turbo}           & 0.46 & 0.21 & 0.63 & 0.00 \\
\textbf{GPT-4o-mini}           & 0.59 & 0.32 & 0.68 & 0.00 \\
\midrule
\textbf{Phi-4-mini}        & \textbf{0.25} & 0.12 & 0.54 & 0.00 \\
\textbf{Qwen-2.5-7B}       & 0.02 & \textbf{0.17} & 0.57 & 0.00 \\
\textbf{Mistral-7B-v0.3}        & 0.08 & 0.03 & 0.59 & 0.00 \\
\textbf{LLaMA-3.1-8B}         & 0.08 & 0.08 & 0.55 & 0.00 \\
\textbf{\modelname{}} & 0.03 & 0.09 & \textbf{0.71} & 0.00 \\
\bottomrule
\end{tabular}}
\vspace{-0.5em}
\label{tab:ood}
\end{table}

% \begin{table}[t]
% \renewcommand{\arraystretch}{1.2}
% \resizebox{7.5cm}{!}{
% \begin{tabular}{ccccc}
% \toprule[1.2pt]
% \multicolumn{1}{l}{\textbf{}} & \multicolumn{1}{l}{\textbf{Degree}} & \textbf{CN} & \multicolumn{1}{l}{\textbf{Diameter}} & \textbf{Edge} \\ \hline
% \textbf{Vicuna-7b}            & 0.14                                & 0.10        & 0.03                                  & 0.07          \\ \hline
% \textbf{Ours}                 & 0.54                                & 0.22        & 0.09                                  & 0.86          \\ \bottomrule[1.2pt]
% \end{tabular}}
% \caption{Zero-shot performance on 4 classical graph reasoning tasks with different node ID representation. \textit{Common Neighbor} is abbreviated as \textit{CN}.}
% \label{tab:node_id}
% \end{table}

% \begin{table}[t]
% \renewcommand{\arraystretch}{1.2}
% \resizebox{4.5cm}{!}{
% \begin{tabular}{ccccc}
% \toprule[1.2pt]
% \textbf{}          & \textbf{BFS} & \textbf{Cycle} & \textbf{CC} & \textbf{EP} \\ \midrule[1.2pt]
% \textbf{Vicuna-7b} & 0.03         & 0.22           & 0.00        & 0.00        \\ \hline
% \textbf{Ours}      & 0.05         & 0.46           & 0.00        & 0.00        \\ \bottomrule[1.2pt]
% \end{tabular}}
% \caption{Zero-shot performance on 4 OOD graph reasoning tasks. \textit{Clustering Coefficient} and \textit{Euler Path} are abbreviated as \textit{CC} and \textit{EP}, respectively.}
% \label{tab:ood}
% \end{table}

\paragraph{Out-of-domain tasks} At last, we evaluate the performance of \modelname{} on four out-of-domain tasks, namely \textit{BFS}, \textit{Cycle}, \textit{Clustering Coefficient}, and \textit{Euler Path}. The zero-shot accuracy is presented in Table~\ref{tab:ood}. It is evident that the performance on out-of-domain tasks is less satisfactory compared to that on in-domain tasks, particularly for \textit{Clustering Coefficient} and \textit{Euler Path}. The results in Table~\ref{tab:ood} indicate that 1) Some graph reasoning tasks like \textit{Euler Path} are still challenging for LLMs, and even powerful proprietary models like GPT-4o-mini can not provide correct answers; 2) Currently the data fed to \modelname{} may be still insufficient, and \modelname{} fails to figure out how to effectively solve unseen graph reasoning tasks, which we leave as future work.

%% file: reasoning.tex
\section{\modelnamestep{}: Towards Graph Reasoning}
\label{sec:step}
Previous experiments have demonstrated that \modelname{} possesses a certain level of graph understanding capability after fine-tuning. However, relying solely on supervision with final answers might limit its graph reasoning capability. In other words, \modelname{} may only have an intuitive understanding of graph data without being able to provide the reasoning steps for each answer. In this part, we further explore how to utilize the intermediate reasoning steps provided by \benchmarkname{} to enhance the graph reasoning capability of LLMs.

\subsection{Methodology}
Different from the training approach in Section~\ref{sec:train}, we incorporate intermediate steps $\mathcal{S}$ as supervised signals into the training data as well to empower LLMs with graph reasoning capability, and construct a new model namely \modelnamestep{}. Based on instruction-tuning, the loss function utilized at this stage can be formulated as:
\begin{equation}
    \mathcal{L}=-\mathbb{E}_{<\mathcal{G},\mathcal{Q},\mathcal{S},\mathcal{A}>\sim \mathcal{D}}{\rm log}P((\mathcal{S},\mathcal{A})|(\mathcal{G},\mathcal{Q});\theta),
\end{equation}
where given a graph $\mathcal{G}$ and corresponding query $\mathcal{Q}$, \modelnamestep{} with parameters $\theta$ are trained to output reasoning steps $\mathcal{S}$ before getting the final answer $\mathcal{A}$.

\begin{figure}[ht]
    \centering
    \begin{minipage}[t]{0.42\textwidth}
    \includegraphics[width = \textwidth]{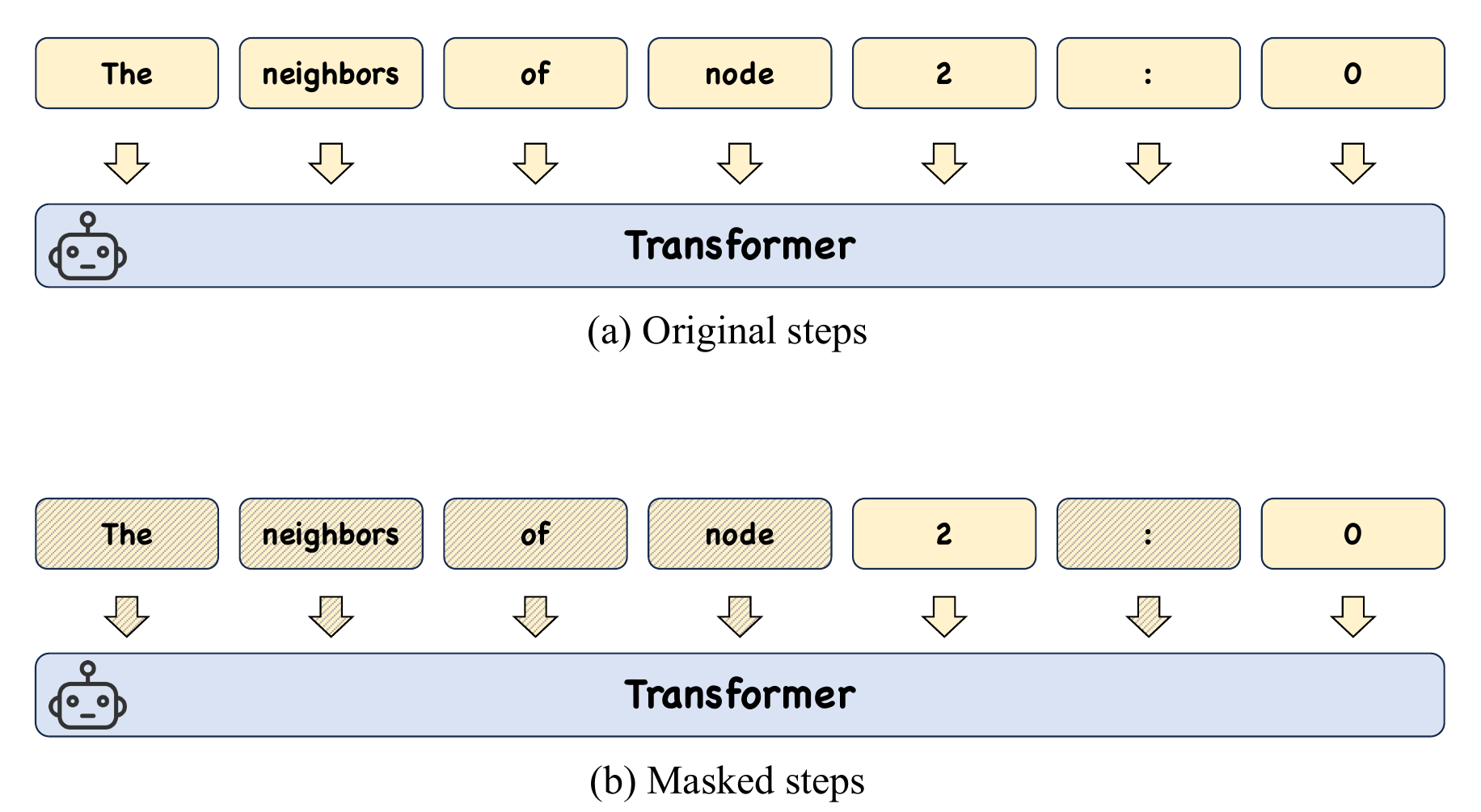}
    \end{minipage}%
    \caption{Toy graph for illustrating the concept of label mask training strategy}
    \label{fig:step_mask}
    \vspace{-1em}
\end{figure}

However, unlike mathematical reasoning tasks \cite{t1}, intermediate steps $\mathcal{S}=\{s_1, s_2, ..., s_T\}$ in graph reasoning tasks may contain numerous repetitive and redundant tokens, as shown in Figure~\ref{fig:step_mask}(a), which could obscure crucial node ID information. Therefore, we employ a label mask training strategy, where the supervised signals corresponding to unimportant tokens are ignored with a certain probability $\gamma$, not contributing to the final loss function, while preserving information relevant to node IDs entirely:
% \begin{equation}
% \mathcal{L} = - \mathbb{E}_{< \mathcal{G}, \mathcal{Q}, \mathcal{S}, \mathcal{A} > \sim \mathcal{D}} 
% \left[ \sum_{t=1}^{T} m_t \cdot \log P\left(s_t, \mathcal{A}) \mid \mathcal{G}, \mathcal{Q}; \theta\right) \right]
% \label{eq:label-mask-loss}
% \end{equation}
% \begin{equation}
% m_t = 
% \begin{cases}
% 1, & \text{if } f(s_t) = 1 \\
% \text{Bernoulli}(1 - \gamma), & \text{if } f(s_t) = 0
% \end{cases}
% \label{eq:mask-definition}
% \end{equation}
\begin{align}
\mathcal{L}
= - \mathbb{E}_{\langle\mathcal{G},\mathcal{Q},\mathcal{S},\mathcal{A}\rangle\sim\mathcal{D}}
\Biggl[ &
\sum_{t=1}^{T} m_t \,\log P\bigl(s_t \mid \mathcal{G},\mathcal{Q},s_{<t};\theta\bigr) \nonumber\\
&\qquad +\; \log P\bigl(\mathcal{A} \mid \mathcal{G},\mathcal{Q},\mathcal{S};\theta\bigr)
\Biggr],
\label{eq:label-mask-loss-aligned}
\end{align}

\begin{equation}
m_t =
\begin{cases}
1, & \text{if } f(s_t)=1 \\
\displaystyle \mathrm{Bernoulli}(1-\gamma), & \text{if } f(s_t)=0
\end{cases}
\label{eq:mask-definition}
\end{equation}
where $m_t\in\{0,1\}$ indicates if the token $s_t$ will be ignored in final loss, and $f(s_t)$ is an indicator function examining if the token $s_t$ contains node ID information. $\gamma$ is a predefined hyperparameter. The benefits of label mask training is two-fold: 1) Through this approach, we can filter out a significant amount of redundant supervised signals, avoiding the overfitting risks on some unimportant tokens; 2) With masking unimportant tokens in the steps, models can better capture the long-distance dependency among node IDs.

\subsection{Main Experimental Results}

\begin{figure}[t]
    \centering
    \begin{minipage}[t]{0.495\textwidth}
    \includegraphics[width = \textwidth]{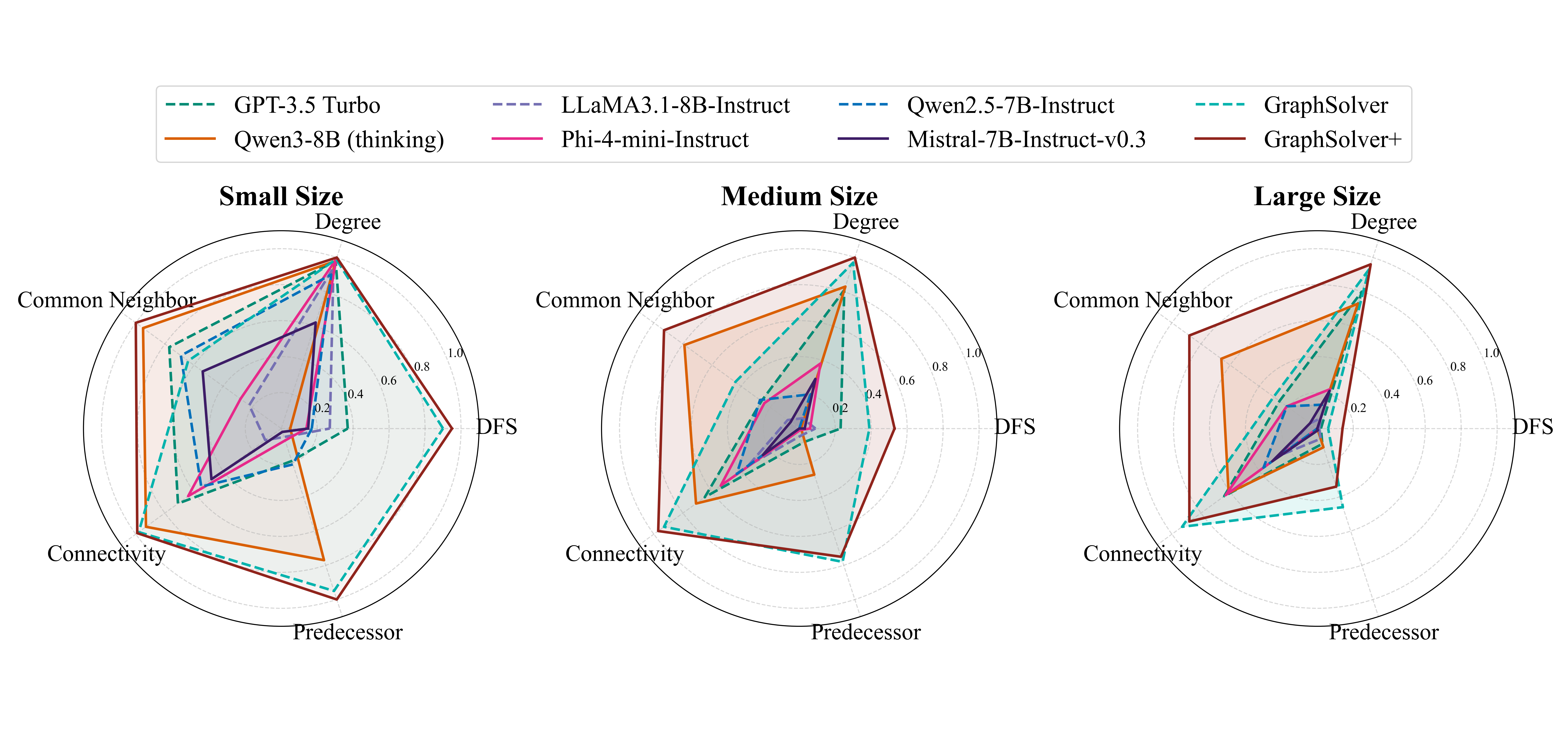}
    \end{minipage}%
    \vspace{-0.5em}
    \caption{Zero-shot CoT performance (Accuracy) comparisons on five classical graph reasoning tasks. Here we enable the thinking mode for Qwen-3B.}
    \label{fig:radar}
\end{figure}

To be concrete, we employ five classical graph reasoning tasks from \benchmarkname{}: \textit{DFS}, \textit{Predecessor}, \textit{Degree}, \textit{Common Neighbor}, and \textit{Connectivity} for training. Each task provides 10,000 training samples, and the token mask probability $\gamma$ is set to 0.8 after tuning. To evaluate the reasoning capacity of \modelnamestep{}, we employ zero-shot CoT techniques that prompt LLMs to generate answers with detailed reasoning steps. As demonstrated in Figure~\ref{fig:radar}, our evaluation across small, medium, and large-scale test sets reveals the following observations:

\begin{itemize}[leftmargin=1em]
    \item \modelnamestep{} significantly outperforms all baseline models across all five tasks, even including reasoning models like Qwen3-8B when intermediate reasoning steps are incorporated into the supervision signals. Notably, in the medium-size DFS task, \modelnamestep{} achieves 53\% accuracy - a sixfold improvement over LLaMA-3.1-8B-Instruct's performance (approximately 9\%).
    \item Additionally, while all models exhibit performance degradation as graph size increases, \modelnamestep{} maintains consistent superiority across all graph sizes, and marginal performance degradation on tasks like \textit{Degree}, \textit{Common Neighbor}, and \textit{Connectivity}. This robustness highlights the potential of \benchmarkname{} in helping models handle increasingly complex graph structures and the generalizability of \modelnamestep{}. 
    \item To our surprise, \modelname{} demonstrates comparable performance with \modelnamestep{} on tasks like \textit{Connectivity} and \textit{Predecessor} when they are asked to address these graph reasoning tasks step by step. However, our further in-depth step evaluation in the following part indicates that \modelname{} actually poses an unsatisfactory reasoning capacity compared to \modelnamestep{}. We leave the corresponding analysis to the next part.
\end{itemize}
 
% In contrast, \modelname{} achieves unsatisfactory results compared with \modelnamestep{}. Unlike the zero-shot setting, we believe that in the zero-shot CoT context, the LLM needs to possess a certain level of graph reasoning ability to correctly produce intermediate step outputs and derive the final answer, which is hard for \modelname{} due to the lack of reasoning steps during fine-tuning.

\begin{table}[t]
\centering
\renewcommand{\arraystretch}{1.3}
\caption{Manual evaluation results (Accuracy) on the reasoning steps of \modelname{} and \modelnamestep{} on medium graphs}
\resizebox{8.4cm}{!}{
\begin{tabular}{lcccccccccc}
\toprule[1.2pt]
& \multicolumn{2}{c}{\textbf{DFS}} & \multicolumn{2}{c}{\textbf{Degree}} & \multicolumn{2}{c}{\textbf{Common Neighbor}} & \multicolumn{2}{c}{\textbf{Connectivity}} & \multicolumn{2}{c}{\textbf{Predecessor}} \\
\cmidrule(lr){2-3}\cmidrule(lr){4-5}\cmidrule(lr){6-7}\cmidrule(lr){8-9}\cmidrule(lr){10-11}
& Answer & Steps & Answer & Steps & Answer & Steps & Answer & Steps & Answer & Steps \\
\midrule
\textbf{GraphSolver} & 0.39 & 0.00 & 0.97 & 0.00 & 0.44 & 0.00 & 0.44 & 0.00 & \textbf{0.78} & 0.00   \\
\textbf{GraphSolver+}         & \textbf{0.53} & \textbf{0.42} & \textbf{1.00} & \textbf{0.98} & \textbf{0.93} & \textbf{0.90} & \textbf{0.97} & \textbf{0.93} & 0.75 & \textbf{0.75}   \\
\bottomrule[1.2pt]
\end{tabular}}
\label{tab:steps-eval}
\end{table}

\subsection{Steps Evaluation} 
To evaluate the reasoning capability of \modelname{} and \modelnamestep{}, we check the correctness of the reasoning steps output by \modelname{} and \modelnamestep{} manually, which contains 100 test samples for each task. Table~\ref{tab:steps-eval} presents the accuracy of the final answer and intermediate steps output by models on five classical graph reasoning tasks under the zero-shot CoT setting, from which we can observe that: When fine-tuning without using intermediate reasoning steps as supervised signals, \modelname{}'s intermediate reasoning step outputs always turn out to be incorrect. For example, in the \textit{Common Neighbor} task, although 44 final answers are correct, none of them has correct intermediate step outputs. The possible reason for this issue is that when only using final answers as supervised signals, the fine-tuned \modelname{} may only develop a shallow understanding of graph data and graph reasoning tasks, unable to provide correct reasoning processes and explanations. In contrast, \modelnamestep{} maintains much higher accuracy in intermediate reasoning steps across all five tasks.

\begin{figure}[t]
    \centering
    \begin{minipage}[t]{0.49\textwidth}
    \includegraphics[width = \textwidth]{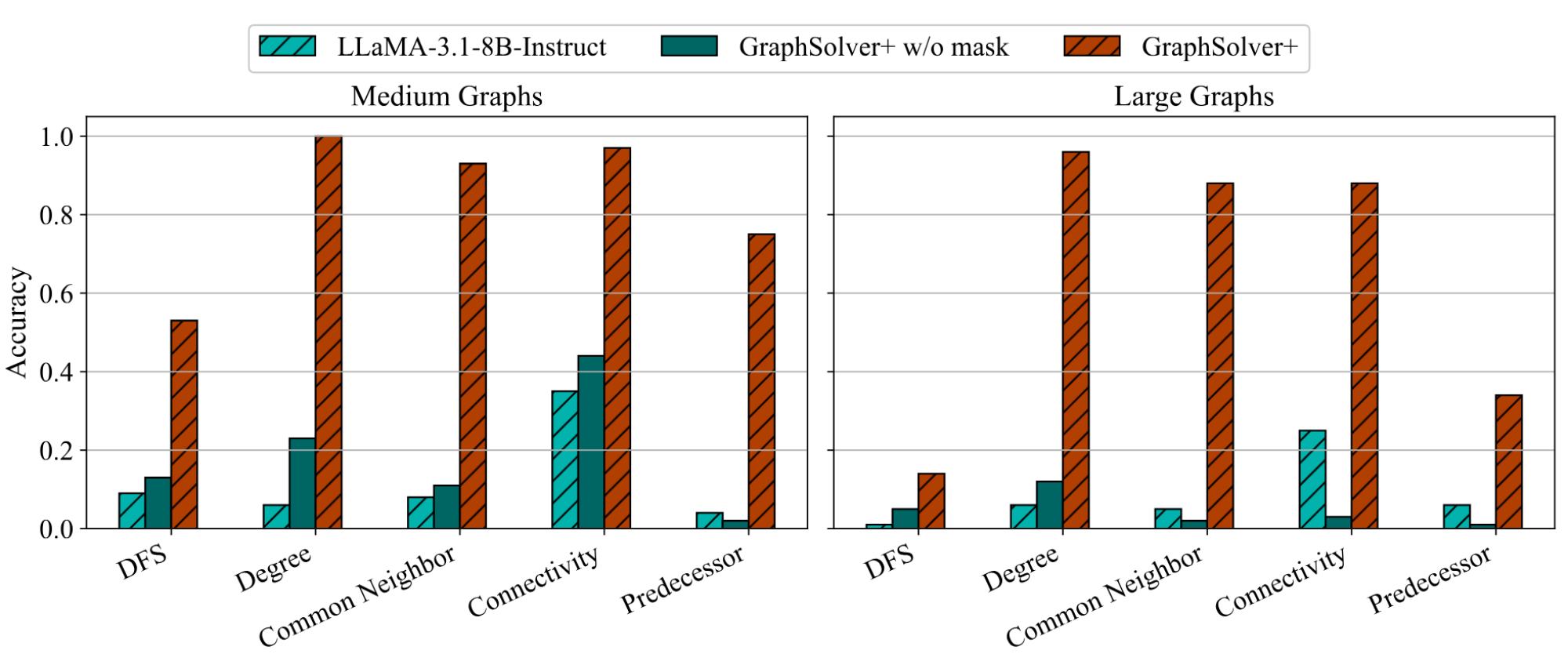}
    \end{minipage}%
    \vspace{-1em}
    \caption{Ablation study of \modelnamestep{}}
    \vspace{-0.5em}
    \label{fig:abaltion}
\end{figure}

\subsection{Ablation Study} 
In this part, we conduct ablation experiments to analyze the effectiveness of our label mask training strategy. Specifically, we remove the label mask strategy under this setting, which means all tokens in the intermediate steps will be treated as supervision signals during fine-tuning. As shown in Figure~\ref{fig:abaltion}, removing the label mask strategy will bring severe performance deterioration on both medium graphs and large graphs. As we mentioned before, without the label mask training strategy, a large number of redundant tokens will be involved in the training process, making it hard for LLMs to capture the key information about the graph structure in the intermediate steps. Similar observations can also be found in math reasoning tasks \cite{mft}.

\subsection{Case Study Analysis} 
\begin{figure}[t]
    \centering
    \includegraphics[width = 0.498\textwidth]{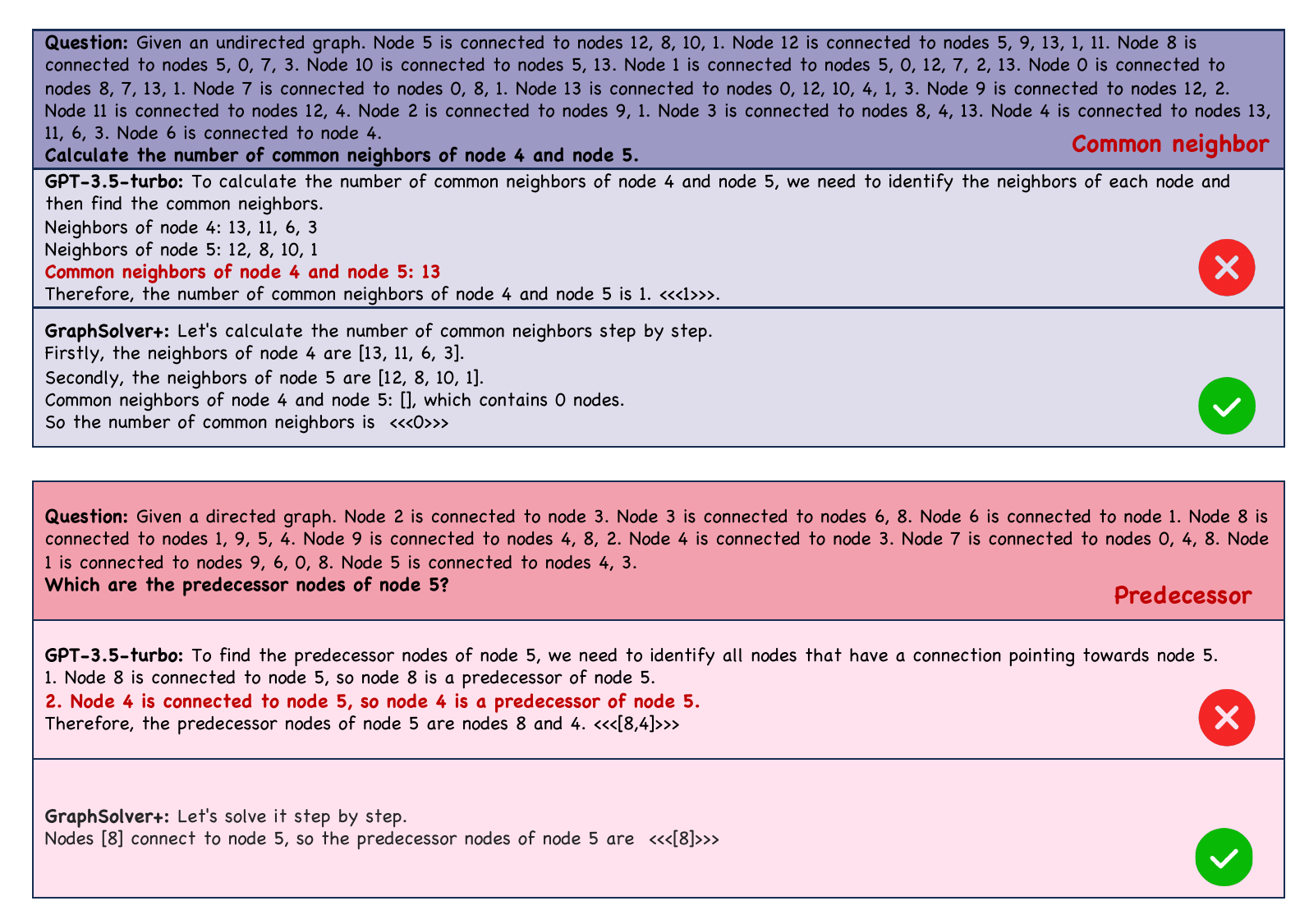}
    \caption{Case study on the output of \modelnamestep{} and GPT-3.5 Turbo}
    \vspace{-0.8pt}
    \label{fig:case}
\end{figure}

In this subsection, we illustrate two detailed cases for better understanding the advantages of \modelnamestep{} in solving graph reasoning tasks. Specifically, we choose \textit{Common neighbors} and \textit{Predecessor} as representative tasks, and demonstrate the results of GPT-3.5 Turbo for comparison. As shown in Figure~\ref{fig:case}., GPT-3.5 Turbo provides incorrect results due to calculation error and misunderstanding about the edges in the directed graph, as highlighted in the Figure. Such incorrect intermediate reasoning steps directly lead to the final wrong results. In contrast, \modelnamestep{} consistently provides correct reasoning steps and outputs the true answer, which demonstrates its superiority in graph reasoning tasks.

% \paragraph{Case study} Here we will elaborate on this observation with detailed case studies for better understanding. As shown in Figure~\ref{fig:case_study}, the red part indicates incorrect output by \modelnamestep{}, and it can be seen that, \modelnamestep{} makes a relatively small mistake on determining the whole neighbor set of node 0, which actually will not affect the search route and final answer.

% \begin{figure}[t]
%     \centering
%     \includegraphics[width = 0.46\textwidth]{figure/case_study.pdf}
%     \caption{Case study on the output of \modelnamestep{} with incorrect reasoning steps.}
%     \label{fig:case_study}
% \end{figure}

%% file: conclusion.tex
\section{Conclusion}
To improve the graph understanding capability and graph reasoning capability of LLMs, we propose a dynamic benchmark called \benchmarkname{} in this paper, which comprehensively provides 21 classic graph reasoning tasks with diverse graph generation pipelines and detailed intermediate reasoning steps. Based on \benchmarkname{}, we construct \modelname{} and \modelnamestep{} through supervised fine-tuning with a novel label mask training strategy. Comprehensive experiments demonstrate that models trained on \benchmarkname{} could pose superior graph understanding capability and reasoning capability. We expect that \benchmarkname{} could serve as a fertile soil for future researchers to enhance LLMs' capabilities in graph data mining.

%% file: limitations.tex
\section*{Limitations and Discussions}
Firstly, we observe that our models still under-perform when facing some complex graph reasoning tasks like \textit{MST} and \textit{PageRank}, and out-of-domain tasks like \textit{Euler Path} and \textit{Clustering Coefficient}. The reasons might be two-fold: 1) The models may struggle with capturing long-range dependencies required by these tasks. Even with intermediate reasoning steps, the \modelnamestep{} may face issues of memory loss or information leakage when reasoning over long or complex paths, especially in larger graphs or tasks requiring detailed numerical computations; 2) The training samples for these tasks may still be insufficient. Currently, the number of training samples for each in-domain task is equivalent, while ignoring the varying difficulty levels for different tasks. In the future, stronger backbone models, memory mechanisms, and training strategies like curriculum learning can be introduced during the training of \modelname{} and \modelnamestep{} to address the above challenges.

Secondly, the current \benchmarkname{} is limited to abstract graph reasoning tasks without specific applications. While these tasks provide a valuable foundation for evaluating and enhancing graph reasoning capabilities, they do not fully capture the complexity of real-world scenarios where graph-based reasoning plays a critical role. Moving forward, one of the key directions for our future work is to explore the development of more application-specific graph reasoning tasks that are directly relevant to real-world problems, such as urban computing and biological network analysis.
% At last, although we report the results of GPT3.5-turbo and GPT4 on \benchmarkname{}, the performance of more LLMs such as Mistral \cite{DBLP:journals/corr/abs-2310-06825}, Qwen \cite{DBLP:journals/corr/abs-2309-16609}, Baichuan2 \cite{DBLP:journals/corr/abs-2309-10305}, etc should be investigated as well to comprehensively examine the LLMs' capability towards graph data, and we will elaborate on this part in the future.

%% file: main.bbl
\begin{thebibliography}{10}

\bibitem{t1}
Hou Z, Lv~X, Lu~R, Zhang J, Li~Y, Yao Z, Li~J, Tang J, Dong Y.
\newblock T1: Advancing language model reasoning through reinforcement learning and inference scaling.
\newblock In: Proceedings of the 42nd International Conference on Machine Learning.
\newblock 2025

\bibitem{wizardmath}
Luo H, Sun Q, Xu~C, Zhao P, Lou J, Tao C, Geng X, Lin Q, Chen S, Tang Y, Zhang D.
\newblock Wizard{M}ath: Empowering mathematical reasoning for large language models via reinforced evol-instruct.
\newblock In: Proceedings of the 13th International Conference on Learning Representations.
\newblock 2025

\bibitem{DBLP:journals/corr/abs-2306-08568}
Luo Z, Xu~C, Zhao P, Sun Q, Geng X, Hu~W, Tao C, Ma~J, Lin Q, Jiang D.
\newblock Wizard{C}oder: Empowering code large language models with evol-instruct.
\newblock In: Proceedings of the 12th International Conference on Learning Representations.
\newblock 2024

\bibitem{mle-bench}
Chan J~S, Chowdhury N, Jaffe O, Aung J, Sherburn D, Mays E, Starace G, Liu K, Maksin L, Patwardhan T, Madry A, Weng L.
\newblock {MLE}-bench: Evaluating machine learning agents on machine learning engineering.
\newblock In: Proceedings of the 13th International Conference on Learning Representations.
\newblock 2025

\bibitem{DBLP:conf/nips/Ouyang0JAWMZASR22}
Ouyang L, Wu~J, Jiang X, Almeida D, Wainwright C~L, Mishkin P, Zhang C, Agarwal S, Slama K, Ray A, Schulman J, Hilton J, Kelton F, Miller L, Simens M, Askell A, Welinder P, Christiano P~F, Leike J, Lowe R.
\newblock Training language models to follow instructions with human feedback.
\newblock In: Advances in Neural Information Processing Systems 35: Annual Conference on Neural Information Processing Systems 2022.
\newblock 2022

\bibitem{CPsyCoun}
Zhang C, Li~R, Tan M, Yang M, Zhu J, Yang D, Zhao J, Ye~G, Li~C, Hu~X.
\newblock C{P}sy{C}oun: {A} report-based multi-turn dialogue reconstruction and evaluation framework for chinese psychological counseling.
\newblock In: Findings of the 62nd Annual Meeting of the Association for Computational Linguistics.
\newblock 2024,  13947--13966

\bibitem{DBLP:conf/nips/AlayracDLMBHLMM22}
Alayrac J, Donahue J, Luc P, Miech A, Barr I, Hasson Y, Lenc K, Mensch A, Millican K, Reynolds M, Ring R, Rutherford E, Cabi S, Han T, Gong Z, Samangooei S, Monteiro M, Menick J~L, Borgeaud S, Brock A, Nematzadeh A, Sharifzadeh S, Binkowski M, Barreira R, Vinyals O, Zisserman A, Simonyan K.
\newblock Flamingo: a visual language model for few-shot learning.
\newblock In: Advances in Neural Information Processing Systems 35: Annual Conference on Neural Information Processing Systems 2022.
\newblock 2022

\bibitem{II-Bench}
Liu Z, Fang F, Feng X, Du~X, Zhang C, Wang N, Bai Y, Zhao Q, Fan L, Gan C, Lin H, Li~J, Ni~Y, Wu~H, Narsupalli Y, Zheng Z, Li~C, Hu~X, Xu~R, Chen X, Yang M, Liu J, Liu R, Huang W, Zhang G, Ni~S.
\newblock I{I}-{B}ench: An image implication understanding benchmark for multimodal large language models.
\newblock In: Advances in Neural Information Processing Systems 38: Annual Conference on Neural Information Processing Systems 2024.
\newblock 2024

\bibitem{DBLP:journals/corr/abs-2304-12995}
Huang R, Li~M, Yang D, Shi J, Chang X, Ye~Z, Wu~Y, Hong Z, Huang J, Liu J, Ren Y, Zou Y, Zhao Z, Watanabe S.
\newblock Audio{GPT}: Understanding and generating speech, music, sound, and talking head.
\newblock In: Proceedings of the 38th {AAAI} Conference on Artificial Intelligence.
\newblock 2024,  23802--23804

\bibitem{xgcn}
Song X, Lian J, Huang H, Luo Z, Zhou W, Lin X, Wu~M, Li~C, Xie X, Jin H.
\newblock x{GCN}: An extreme graph convolutional network for large-scale social link prediction.
\newblock In: Proceedings of the ACM Web Conference 2023.
\newblock 2023

\bibitem{fraud}
Cheng D, Zou Y, Xiang S, Jiang C.
\newblock Graph neural networks for financial fraud detection: a review.
\newblock Frontiers Comput. Sci., 2025, 19(9): 199609

\bibitem{bio}
Zhang X, Lei X.
\newblock Predicting mirna-drug interactions via dual-channel network based on {TCN} and bilstm.
\newblock Frontiers Comput. Sci., 2025, 19(5): 195905

\bibitem{he2024integrating}
He~H, Chen G, Chen C~Y~C.
\newblock Integrating sequence and graph information for enhanced drug-target affinity prediction.
\newblock Science China Information Sciences, 2024, 67(2): 129101

\bibitem{anonymous2024dyval}
Zhu K, Chen J, Wang J, Gong N~Z, Yang D, Xie X.
\newblock Dy{V}al: Graph-informed dynamic evaluation of large language models.
\newblock In: Proceedings of the 12th International Conference on Learning Representations.
\newblock 2024

\bibitem{graphsilo}
Peng M, Chen N, Suo Z, Li~J.
\newblock Rewarding graph reasoning process makes llms more generalized reasoners.
\newblock In: Proceedings of the 31st ACM SIGKDD Conference on Knowledge Discovery and Data Mining.
\newblock 2025,  2257--2268

\bibitem{GraphPile}
Zhang Q, Chen N, Li~Z, Peng M, Tang J, Li~J.
\newblock Improving llms' generalized reasoning abilities by graph problems.
\newblock In: Proceedings of the 2nd Conference on Language Modeling.
\newblock 2025

\bibitem{DBLP:journals/corr/abs-2305-15066}
Guo J, Du~L, Liu H.
\newblock G{PT}4{G}raph: Can large language models understand graph structured data? {A}n empirical evaluation and benchmarking.
\newblock CoRR, 2023, abs/2305.15066

\bibitem{wang2023can}
Wang H, Feng S, He~T, Tan Z, Han X, Tsvetkov Y.
\newblock Can language models solve graph problems in natural language?
\newblock In: Advances in Neural Information Processing Systems 36: Annual Conference on Neural Information Processing Systems 2023.
\newblock 2023

\bibitem{grapharena}
Tang J, Zhang Q, Li~Y, Chen N, Li~J.
\newblock Graph{A}rena: Evaluating and exploring large language models on graph computation.
\newblock In: Proceedings of the 13th International Conference on Learning Representations.
\newblock 2025

\bibitem{fatemi2023talk}
Fatemi B, Halcrow J, Perozzi B.
\newblock Talk like a graph: Encoding graphs for large language models.
\newblock In: Proceedings of the 12th International Conference on Learning Representations.
\newblock 2024

\bibitem{chai2023graphllm}
Chai Z, Zhang T, Wu~L, Han K, Hu~X, Huang X, Yang Y.
\newblock Graph{LLM}: Boosting graph reasoning ability of large language model.
\newblock CoRR, 2023, abs/2310.05845

\bibitem{tang2023graphgpt}
Tang J, Yang Y, Wei W, Shi L, Su~L, Cheng S, Yin D, Huang C.
\newblock Graph{GPT}: Graph instruction tuning for large language models.
\newblock In: Proceedings of the 47th International {ACM} {SIGIR} Conference on Research and Development in Information Retrieval.
\newblock 2024,  491--500

\bibitem{graphwiz}
Chen N, Li~Y, Tang J, Li~J.
\newblock Graph{W}iz: An instruction-following language model for graph computational problems.
\newblock In: Proceedings of the 30th {ACM} {SIGKDD} Conference on Knowledge Discovery and Data Mining.
\newblock 2024,  353--364

\bibitem{DBLP:conf/nips/KojimaGRMI22}
Kojima T, Gu~S~S, Reid M, Matsuo Y, Iwasawa Y.
\newblock Large language models are zero-shot reasoners.
\newblock In: Advances in Neural Information Processing Systems 35: Annual Conference on Neural Information Processing Systems 2022.
\newblock 2022

\bibitem{DBLP:conf/nips/Wei0SBIXCLZ22}
Wei J, Wang X, Schuurmans D, Bosma M, Ichter B, Xia F, Chi E~H, Le~Q~V, Zhou D.
\newblock Chain-of-thought prompting elicits reasoning in large language models.
\newblock In: Advances in Neural Information Processing Systems 35: Annual Conference on Neural Information Processing Systems 2022.
\newblock 2022

\bibitem{graphomni}
Xu~H, Jian X, Zhao X, Pang W, Zhang C, Wang S, Zhang Q, Monteiro J, Sun Q, Yu~T.
\newblock Graph{O}mni: {A} comprehensive and extendable benchmark framework for large language models on graph-theoretic tasks.
\newblock CoRR, 2025, abs/2504.12764

\bibitem{GCoder}
Zhang Q, Hong X, Tang J, Chen N, Li~Y, Li~W, Tang J, Li~J.
\newblock G{C}oder: Improving large language model for generalized graph problem solving.
\newblock In: Proceedings of the 34th ACM International Conference on Information and Knowledge Management.
\newblock 2025,  4149--4159

\bibitem{Graphforge}
Wang R, Liang S, Chen Q, Zhang J, Qin K.
\newblock Graph{T}ool-{I}nstruction: Revolutionizing graph reasoning in llms through decomposed subtask instruction.
\newblock In: Proceedings of the 31st {ACM} {SIGKDD} Conference on Knowledge Discovery and Data Mining, V.1.
\newblock 2025,  1492--1503

\bibitem{wizardlm}
Xu~C, Sun Q, Zheng K, Geng X, Zhao P, Feng J, Tao C, Lin Q, Jiang D.
\newblock Wizard{LM}: Empowering large pre-trained language models to follow complex instructions.
\newblock In: Proceedings of the 12th International Conference on Learning Representations.
\newblock 2024

\bibitem{bigcodebench}
Zhuo T~Y, Vu~M~C, Chim J, Hu~H, Yu~W, Widyasari R, Yusuf I~N~B, Zhan H, He~J, Paul I, Brunner S, Gong C, Hoang T, Zebaze A~R, Hong X, Li~W, Kaddour J, Xu~M, Zhang Z, Yadav P, Jain N, Gu~A, Cheng Z, Liu J, Liu Q, Wang Z, Lo~D, Hui B, Muennighoff N, Fried D, Du~X, Vries d~H, Werra v~L.
\newblock Big{C}ode{B}ench: Benchmarking code generation with diverse function calls and complex instructions.
\newblock In: Proceedings of the 13th International Conference on Learning Representations.
\newblock 2025

\bibitem{bao2023tallrec}
Bao K, Zhang J, Zhang Y, Wang W, Feng F, He~X.
\newblock {TALLR}ec: An effective and efficient tuning framework to align large language model with recommendation.
\newblock In: Proceedings of the 17th ACM Conference on Recommender Systems.
\newblock 2023,  1007–1014

\bibitem{li2023ecomgpt}
Li~Y, Ma~S, Wang X, Huang S, Jiang C, Zheng H, Xie P, Huang F, Jiang Y.
\newblock Ecom{GPT}: Instruction-tuning large language models with chain-of-task tasks for e-commerce.
\newblock In: Proceedings of the 38th {AAAI} Conference on Artificial Intelligence.
\newblock 2024,  18582--18590

\bibitem{HealthLLM}
Kim Y, Xu~X, McDuff D, Breazeal C, Park H~W.
\newblock Health-{LLM}: Large language models for health prediction via wearable sensor data.
\newblock In: Proceedings of the 5th Conference on Health, Inference, and Learning.
\newblock 2024,  522--539

\bibitem{mmedc}
Qiu P, Wu~C, Zhang X, Lin W, Wang H, Zhang Y, Wang Y, Xie W.
\newblock Towards building multilingual language model for medicine.
\newblock Nature Communications, 2024, 15(1): 8384

\bibitem{yue2024lawllm}
Yue S, Liu S, Zhou Y, Shen C, Wang S, Xiao Y, Li~B, Song Y, Shen X, Chen W, others .
\newblock Law{LLM}: Intelligent legal system with legal reasoning and verifiable retrieval.
\newblock In: Proceedings of the 29th International Conference on Database Systems for Advanced Applications.
\newblock 2024,  304--321

\bibitem{doi:10.1126/science.286.5439.509}
Barabási A~L, Albert R.
\newblock Emergence of scaling in random networks.
\newblock Science, 1999, 286(5439): 509--512

\bibitem{watts1998collective}
Watts D~J, Strogatz S~H.
\newblock Collective dynamics of ‘small-world’ networks.
\newblock Nature, 1998, 393(6684): 440--442

\bibitem{DBLP:conf/www/JehW03}
Jeh G, Widom J.
\newblock Scaling personalized web search.
\newblock In: Proceedings of the 12th International World Wide Web Conference.
\newblock 2003,  271--279

\bibitem{nlgift}
Zhang Y, Wang H, Feng S, Tan Z, Han X, He~T, Tsvetkov Y.
\newblock Can {LLM} graph reasoning generalize beyond pattern memorization?
\newblock In: Findings of the 2024 Conference on Empirical Methods in Natural Language Processing.
\newblock 2024,  2289--2305

\bibitem{DBLP:conf/iclr/HuSWALWWC22}
Hu~E~J, Shen Y, Wallis P, Allen{-}Zhu Z, Li~Y, Wang S, Wang L, Chen W.
\newblock Lo{RA}: Low-rank adaptation of large language models.
\newblock In: Proceedings of the 10th International Conference on Learning Representations.
\newblock 2022

\bibitem{DBLP:journals/corr/KingmaB14}
Kingma D~P, Ba~J.
\newblock Adam: {A} method for stochastic optimization.
\newblock In: Proceedings of the 3rd International Conference on Learning Representations.
\newblock 2015

\bibitem{llamafactory}
Zheng Y, Zhang R, Zhang J, Ye~Y, Luo Z.
\newblock {L}lama{F}actory: Unified efficient fine-tuning of 100+ language models.
\newblock In: Proceedings of the 62nd Annual Meeting of the Association for Computational Linguistics.
\newblock 2024,  400--410

\bibitem{vllm}
Kwon W, Li~Z, Zhuang S, Sheng Y, Zheng L, Yu~C~H, Gonzalez J~E, Zhang H, Stoica I.
\newblock Efficient memory management for large language model serving with pagedattention.
\newblock In: Proceedings of the ACM SIGOPS 29th Symposium on Operating Systems Principles.
\newblock 2023

\bibitem{gcn}
Kipf T~N, Welling M.
\newblock Semi-supervised classification with graph convolutional networks.
\newblock In: Proceedings of the 5th International Conference on Learning Representations.
\newblock 2017

\bibitem{graphsage}
Hamilton W~L, Ying Z, Leskovec J.
\newblock Inductive representation learning on large graphs.
\newblock In: Advances in Neural Information Processing Systems 30: Annual Conference on Neural Information Processing Systems 2017.
\newblock 2017

\bibitem{mft}
Chen C, Wang X, Lin T~E, Lv~A, Wu~Y, Gao X, Wen J~R, Yan R, Li~Y.
\newblock Masked thought: Simply masking partial reasoning steps can improve mathematical reasoning learning of language models.
\newblock In: Proceedings of the 62nd Annual Meeting of the Association for Computational Linguistics.
\newblock 2024,  5872--5900

\end{thebibliography}
